\documentclass[a4paper,fleqn]{cas-dc}

\usepackage[authoryear,round,sort]{natbib}

\usepackage{siunitx}
\usepackage{bm}
\usepackage{subcaption}
\usepackage{algorithm}
\usepackage{algpseudocode}

\providecommand{\mbf}[1]{\mathbf{#1}}
\providecommand{\mbs}[1]{\bm{#1}}
\providecommand{\trans}{\top}
\DeclareMathOperator*{\argmin}{arg\,min}

\graphicspath{{./}}

\setkeys{Gin}{keepaspectratio, height=0.92\textheight}

\ExplSyntaxOn
\cs_set:Npn \__first_footerline:
{
  \group_begin: \small \sffamily \__short_authors: \group_end:
}
\ExplSyntaxOff

\begin{document}
\let\WriteBookmarks\relax
\def\floatpagepagefraction{1}
\def\textpagefraction{.001}

\shorttitle{LARK: Low-Cost Occlusion-Resilient Tracking for Image-Guided Surgery}
\shortauthors{G. Sideris et~al.}

\title[mode = title]{LARK: A Low-Cost, Accurate, Occlusion-Resilient, Kalman Filter-Assisted Tracking System for Image-Guided Surgery}

\author[1]{George Sideris}[orcid=https://orcid.org/0009-0003-1355-6606]
\cormark[1]
\fnmark[1]
\ead{george.sideris@mail.mcgill.ca}
\credit{Conceptualization, Methodology, Software, Validation, Formal analysis, Investigation, Data curation, Visualization, Writing -- original draft, Writing -- review \& editing, Project administration}

\author[1]{Justin Cree}
\fnmark[1]
\credit{Conceptualization, Methodology, Software, Validation, Formal analysis, Investigation, Data curation, Visualization, Writing -- review \& editing}

\author[1]{Andrew Stirling}
\credit{Methodology, Software, Writing -- review \& editing}

\author[1]{Mamadou Ly}
\credit{Methodology, Writing -- review \& editing}

\author[2]{\'Etienne L\'eger}
\credit{Conceptualization, Investigation, Supervision, Writing -- review \& editing}

\author[2]{D. Louis Collins}
\credit{Conceptualization, Funding acquisition, Resources, Supervision, Writing -- review \& editing}

\affiliation[1]{organization={Department of Mechanical Engineering, McGill University},
            city={Montreal},
            country={Canada}}

\affiliation[2]{organization={Montreal Neurological Institute, McGill University},
            city={Montreal},
            country={Canada}}

\cortext[1]{Corresponding author.}
\fntext[1]{These authors contributed equally to this work.}
\nonumnote{The supplementary document, available in the arXiv ancillary files, contains the bill of materials and technical drawings (Supplementary Figures S1--S25). A supplementary video demonstrates the LARK system, experimental setups, and key results: \url{https://youtu.be/ApZ8q9DjB-4}.}

\begin{abstract}
Image-guided surgery (IGS) depends on accurate tracking of surgical instruments to provide real-time navigation relative to anatomical structures. Commercial stereo infrared trackers are accurate but prone to occlusion and cost-prohibitive for many settings. This work presents LARK, a multi-camera optical tracking system using commodity RGB hardware and multi-view redundancy and fusion. We develop and evaluate two complete tracking methods: multi-view monocular pose fusion and multi-view triangulation. Both methods are assessed under varying occlusion levels using a precision-machined grid and an anatomical head phantom, and compared against a gold-standard stereo infrared system. With five cameras and adaptive Kalman filtering, LARK achieves median target registration errors of 0.64 mm for point localization with triangulation and 0.73 mm for trajectory tracking with pose fusion on the machined grid. Camera-subset experiments show graceful degradation in adaptive pose-fusion accuracy as fewer views remain available. With tracking hardware costing under \$1,000 USD, LARK provides a low-cost platform for image-guided surgery research. Hardware designs and software are publicly available at \url{https://nist.mni.mcgill.ca/software/}, and datasets at \url{https://nist.mni.mcgill.ca/data/}.
\end{abstract}


\maketitle

\section{Introduction}

Image-guided surgery (IGS) has increasingly facilitated minimally invasive procedures, including renal, orthognathic, spinal, and brain surgeries, by providing real-time navigation tools that enhance accuracy and safety \citep{Jackson2020, VandeBempt2018, Gueziri2021, Wallace2019}. IGS systems track the position of surgical instruments and align them with a three-dimensional model of the patient’s anatomy, created from pre- and/or intra-operative imaging. This information is visualized on a screen or head-mounted device \citep{Asadi31122024}, allowing surgeons to accurately position their instruments relative to critical anatomical structures that are otherwise hidden from view. In the context of image-guided neurosurgery (IGNS), this feedback is crucial for planning and executing minimally invasive procedures, such as tumor resections and biopsies, where visibility is often limited \citep{Taleb2023}.

Tracking in IGNS is commonly achieved using either electromagnetic or optical systems, each with distinct advantages and limitations. Electromagnetic tracking systems typically use an external field generator and miniature sensors attached to or embedded in the surgical instrument. The sensors measure the generated field, allowing the system to estimate their position and orientation. While these systems do not require a direct line of sight, they are susceptible to interference from metal objects in the operating room (OR), which can significantly compromise their reliability and accuracy \citep{Sorriento2020, Jin2022, Poulin2002}.

Optical tracking systems have become the dominant technology in IGNS due to their high accuracy and straightforward integration in the OR. These systems typically use stereo infrared cameras and computer-vision algorithms to detect retro-reflective markers affixed to surgical instruments and to a static reference affixed to the patient. However, they are highly susceptible to occlusion: when surgeons or staff obstruct the camera’s view, tracking is disrupted entirely, often freezing the navigation display and halting guidance until the view is restored \citep{Mehbodniya2019, Han2023}. In addition, the systems are often bulky, adding to the already crowded surgical workspace \citep{Sorriento2020, Lin2023}. Their cost, typically \$8,000 - \$25,000 USD for the tracking module alone, further limits adoption in resource-constrained settings \citep{Leger2021}.

In this paper, we address these limitations through a low-cost multi-camera optical tracking system mounted on a surgical light. The system addresses two distinct occlusion challenges: camera-level occlusion, where entire camera viewpoints are blocked by surgical staff or equipment, and marker-level occlusion, where individual markers are obscured from specific viewpoints. The hardware design, a ring-shaped fixture with distributed cameras, minimises camera-level occlusion by ensuring multiple lines of sight to the surgical field. To handle marker-level occlusion, we implement and compare two pose estimation methods with adaptive Kalman filtering: multi-view monocular pose fusion and multi-view triangulation. We evaluate both approaches using a precision-machined grid and an anatomical head phantom.

The contributions of this work are threefold. First, we present LARK, a low-cost multi-camera optical tracking system using commodity RGB hardware. The surgical-light-mounted ring fixture provides multiple views without external tracking towers. With tracking hardware costing under \$1{,}000~USD, LARK achieves sub-millimeter target registration error on the machined grid with five cameras and adaptive filtering (0.64~mm point localization with triangulation, 0.73~mm trajectory tracking with pose fusion).

Second, we develop a complete tracking methodology comprising a unified calibration pipeline and two pose estimation methods. The calibration pipeline includes multi-camera extrinsic calibration with global bundle adjustment and instrument calibration through bundle-adjusted marker corner refinement. Building on this shared foundation, we implement and systematically compare two pose estimation methods: multi-view monocular pose fusion, which computes independent transformations from each camera view and combines them through adaptive Kalman filtering, and multi-view triangulation, which reconstructs 3D marker positions across cameras before estimating pose. Our comparison reveals complementary performance characteristics: with five cameras and adaptive filtering, monocular pose fusion achieves lower trajectory error on the grid, while triangulation achieves lower point and trajectory errors on the head phantom and is less affected by the illustrated poor single-marker view.

Third, we provide a comprehensive open-source benchmark for surgical tracking system evaluation. Using a precision-machined grid and an anatomical head phantom, we compare both methods across camera subsets representing different levels and patterns of camera unavailability. All hardware designs, bill of materials, software, and 5,000 multi-view tracking videos are publicly released to establish an accessible reference platform for IGS research.

The remainder of this paper is organized as follows. Section~\ref{sec:related_work} reviews state-of-the-art optical tracking systems for neuronavigation and optical tracking using planar fiducial markers. Section~\ref{sec:methods} describes the system hardware, calibration, and pose estimation pipeline. Section~\ref{sec:experiments} presents the experimental protocol. Section~\ref{sec:results} presents the quantitative results. Section~\ref{sec:discussion} discusses the system's performance, limitations and clinical implications, and Section~\ref{sec:conclusion} concludes.

\section{Related Work}\label{sec:related_work}

\subsection{State-of-the-Art Optical Tracking Systems}

Contemporary state-of-the-art neuronavigation systems, such as Atracsys' fusionTrack 500 \citep{Atracsys2024}, NDI’s Polaris Vega \citep{NDI2024}, Stryker's NAV3i \citep{Stryker2024}, and Claronav's Navient \citep{Claronav2024}, rely on a single stereo infrared camera for surgical instrument tracking. These systems achieve tracking through retro-reflective markers mounted on the surgical instrument and a reference fixed to the patient's head. \citet{Fattori2021} demonstrated that NDI's Polaris can achieve sub-millimeter accuracy with latency below 20 milliseconds from as far as 2.4 meters away. However, these systems are expensive, large, and prone to occlusion, which disrupts tracking. To expand the workspace of these systems, \citet{Pfeiffer2016} employed multiple Polaris cameras with data fusion, while \citet{Marinetto2018} utilized multiple OptiTrack cameras, a more affordable alternative typically used in motion capture, but not approved for surgical use. Nevertheless, both approaches remain costly.

\subsection{Optical Tracking Using Planar Fiducial Markers}

In response to these limitations, several efforts have focused on using low-cost RGB cameras for instrument tracking through planar fiducial marker systems, with ArUco \citep{Garrido2014} and AprilTag \citep{Olson2011} being among the most popular due to their accessibility and ease of use. However, studies by \citet{Asselin2018} and \citet{Leger2021} indicate that tracking a single planar marker with an off-the-shelf stereo RGB-Depth camera does not provide the accuracy required for surgical navigation. More recent efforts have explored using multiple planar markers attached to the instrument and/or multiple RGB cameras, either for multi-view monocular tracking with intelligent pose fusion or for stereo tracking.

\subsubsection{Multi-Marker Instrument Tracking}

Research has been done on tracking configurations of multiple fiducial markers to improve object tracking from different perspectives. \citet{Wu2017} proposed DodecaPen, which tracks a dodecahedron of markers using a single camera positioned near the object. However, their work does not consider the larger distances typically observed in OR settings and does not address occlusion at the camera or marker level. \citet{Garcia2023} introduced DodUcoPose, an improvement on the DodecaPen system, designed to track objects using a configuration of multiple pentagonal markers arranged in a dodecahedron. However, DodUcoPose's tracking accuracy evaluations remain confined to relatively short object-to-camera distances, and the method is based on a singular camera viewpoint, making it susceptible to camera-level occlusion. More recently, \citet{Moviglia2026} paired a single near-infrared camera with compact retroreflective dodecahedral fiducials, achieving 0.44~mm translational accuracy at 30--70~cm working distance, though from static position estimates, at short range, and again from a single viewpoint.

\subsubsection{Multi-View RGB Tracking}

Pose estimation using multiple RGB cameras requires precise calibration. For monocular tracking, intrinsic calibration is necessary for each individual camera, while multi-view triangulation demands both intrinsic and extrinsic calibration for all cameras involved. Following calibration, data fusion is typically employed to enhance localization accuracy, particularly in scenarios involving partial occlusion or degraded views from one or more cameras. \citet{Wang2015} proposed a fast non-iterative multi-camera calibration algorithm for determining the extrinsic parameters of a set of monocular cameras, although their approach does not account for intrinsic camera parameters. \citet{Vaccarella2013} introduced an unscented Kalman filter-based sensor fusion method to mitigate marker occlusions and electromagnetic distortions in surgical navigation using an electromagnetic tracking system. \citet{Popescu2020} proposed an adaptive Kalman filter method for fusing planar marker poses from multiple cameras, but their approach assumes static dynamics and their results were only tested in simulation. \citet{Tsui2024} achieved a root-mean-square error of 2 mm when tracking a planar marker using a custom stereo camera pair with a wider baseline than typical off-the-shelf systems. However, their evaluation setup did not mimic OR conditions, as the marker was only tracked within a single plane of movement and at distances typically less than a meter away from the camera pair.

\subsubsection{Hybrid Multi-Camera, Multi-Marker Systems}

\citet{Chen2023} mounted multiple RGB camera modules on a surgical light and tracked surgical instruments using planar fiducial markers. Their system uses multi-camera calibration and least-squares fusion and reported mean point and trajectory errors of 2.01~mm and 2.37~mm, respectively. However, the evaluation was limited to millimeter-level accuracy rather than sub-millimeter accuracy at neurosurgical working distances, and the system design and software were not released publicly.

In this work, we pursue a similar overhead, light-mounted form factor but target sub-millimeter localization accuracy using only commodity hardware. We also consider explicit loss of individual camera views and we present two complementary estimation modes, multi-view monocular pose fusion and multi-view triangulation. The full mechanical fixture, calibration, tracking and registration software, and bill of materials are made available for reproduction.

\subsection{Marker-Less Instrument Tracking}
Marker-less multi-view pose estimation has recently emerged as an alternative that leaves instruments unmodified. \citet{Hein2025} evaluate learning-based methods on a multi-camera dataset of ex-vivo spine surgeries, reporting 1.01~mm instrument position error with five static cameras under optimal in-domain conditions, degrading for a second instrument (2.79~mm) and in an unseen operating room. Accuracy is reported as pose error against a marker-based tracker whose own calibration accuracy the errors approach, rather than as registration-based target error in a patient reference frame, and each instrument requires its own model and training data. Marker-based tracking therefore remains the reference standard where sub-millimeter, instrument-agnostic accuracy is required.

This work compares pose fusion and triangulation on shared hardware and recordings under controlled camera-subset removal, and releases the hardware designs, software, and data as a benchmark for further evaluation.

\section{System Overview and Methods}\label{sec:methods}

\subsection{System Overview}

\begin{figure*}[width=\textwidth,pos=!tbp]
    \centering
    \includegraphics[width=\linewidth]{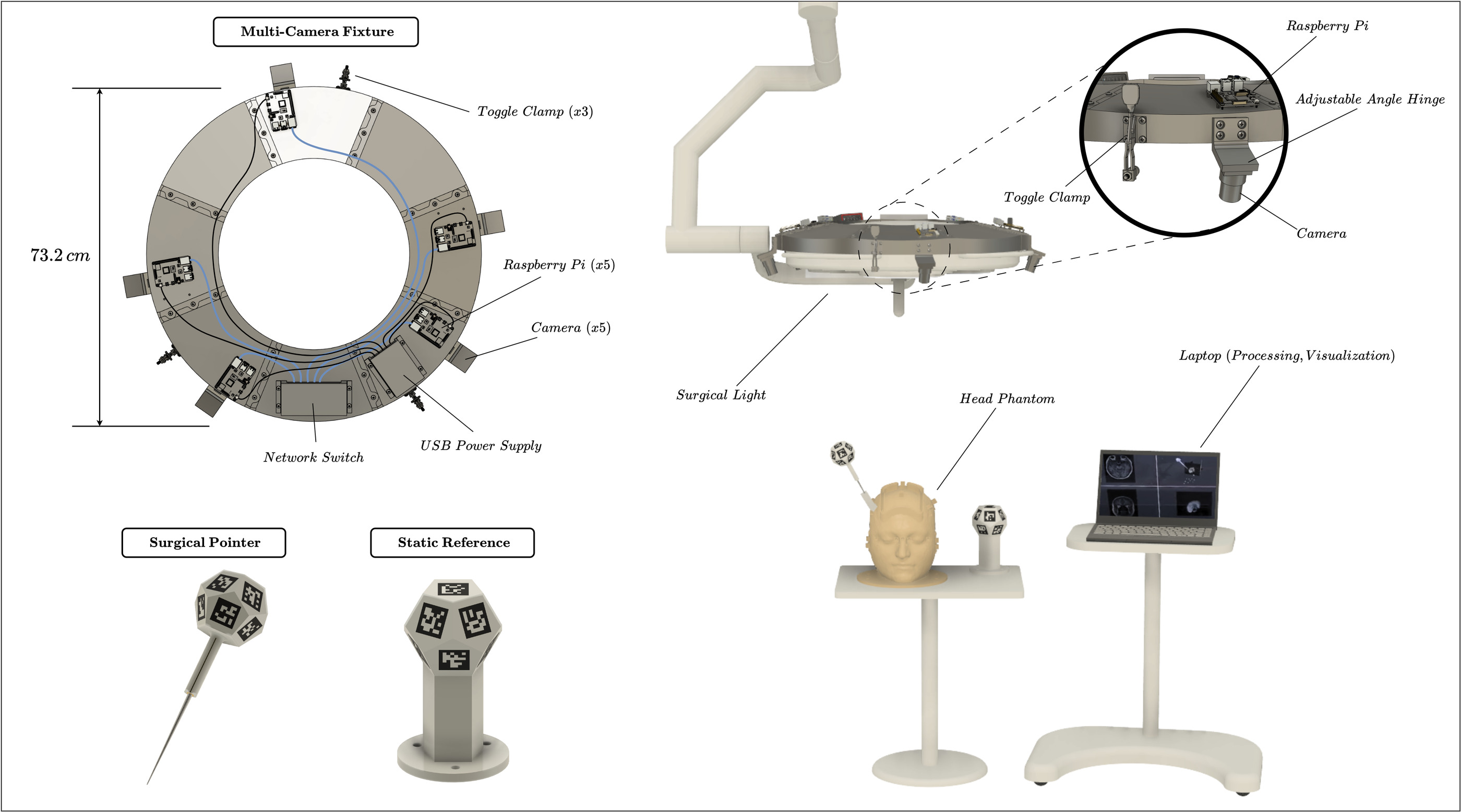}
    \caption{Annotated schematic of the proposed system. The system is composed of a multi-camera fixture (top left), a surgical pointer and a static reference (bottom left). The fixture is attached to a surgical light to track the pointer relative to the reference (right). Typical surgical light geometries consist of an articulated arm whose base is mounted to the OR ceiling. The position of the light can be freely manipulated to illuminate the surgical area, and by doing so, ensure that the cameras maintain line of sight with the anatomy of interest.}
    \label{fig:overview}
  \end{figure*}

\begin{figure}[pos=!tbp]
    \centering
    \includegraphics[width=\linewidth]{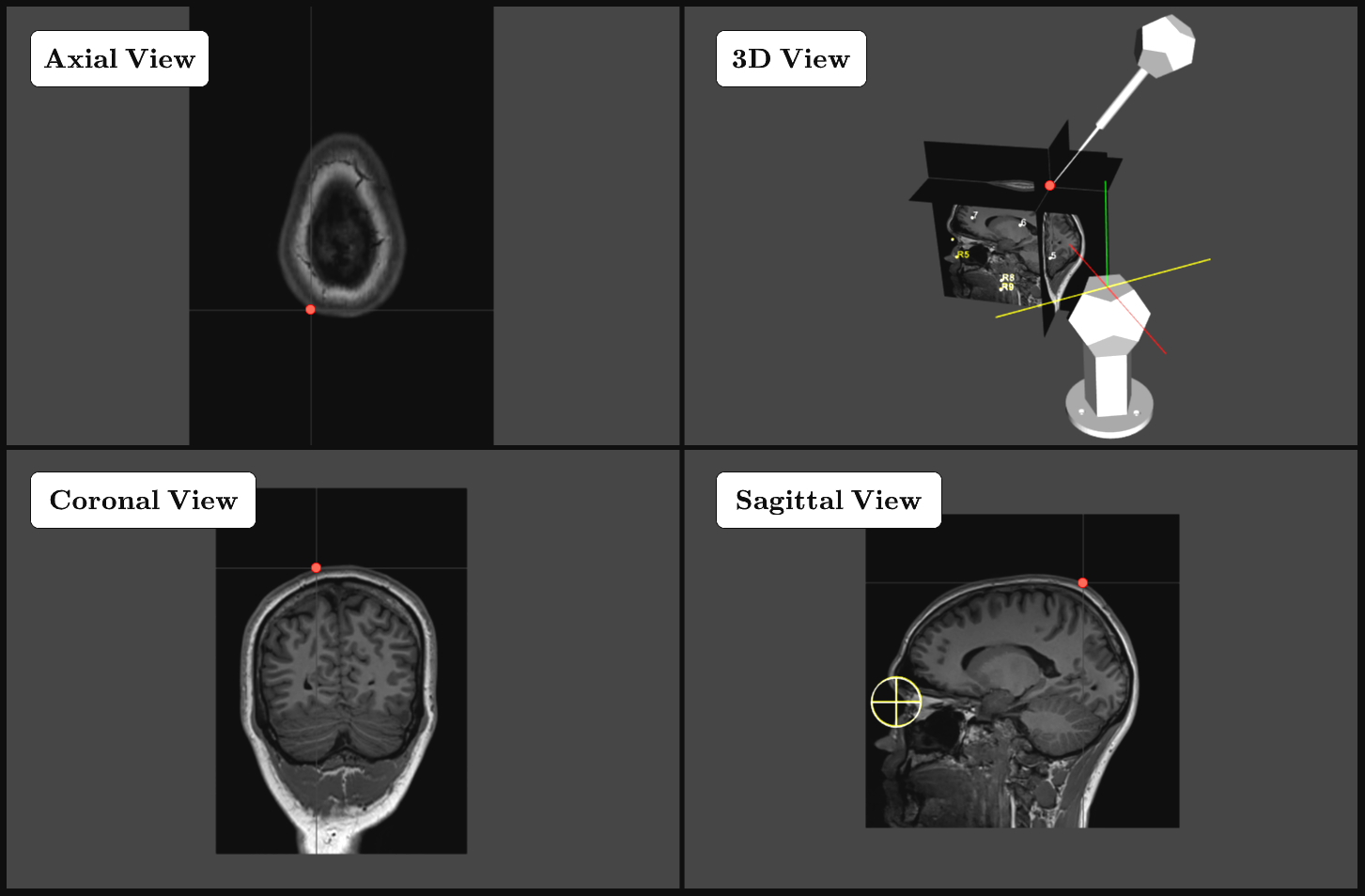}
    \caption{Real-time surgical navigation interface using the IBIS platform~\citep{Drouin2017}. The tracked pointer position (red dot) is displayed on axial, coronal, and sagittal MRI views, with a 3D rendering showing the pointer, reference dodecahedron, and patient anatomy.}
    \label{fig:ibis_visualization}
  \end{figure}

The proposed system consists of a ring-shaped fixture equipped with multiple RGB cameras, secured atop a surgical light to maintain a clear line of sight to the surgical field (see Fig.~\ref{fig:overview}). A surgical pointer instrument and a static reference are each equipped with planar fiducial markers, enabling tracking of the pointer with respect to the reference. Because the tracked entity is the marker assembly rather than the instrument itself, any rigid instrument with a well-defined working point can be tracked after a one-time pivot calibration, without per-instrument models or training. The system is capable of two tracking methods that are evaluated and compared here, both employing adaptive Kalman filtering: one based on fusion of multi-view monocular pose estimates, and one based on multi-view triangulation. Real-time visualization and navigation are achieved through integration with the IBIS neurosurgical navigation platform~\citep{Drouin2017}, which displays the tracked pointer position on patient imaging data. Figure~\ref{fig:ibis_visualization} demonstrates the system's real-time tracking capability, showing the pointer position overlaid on multi-planar MRI views. The supplementary video demonstrates the assembled system, both experimental setups, and real-time tracking with the navigation display.

\subsection{Hardware Design}
\subsubsection{Multi-Camera Fixture}

The primary design objective of our system is to maintain line of sight with the surgical instruments and patient without disrupting surgical workflow and while preserving the OR workspace. To achieve this, a ring-shaped fixture is designed and built to accommodate the attachment of multiple cameras (in this work, we use 5) and be placed on surgical lights. The fixture is composed of eight 3D-printed interlocking segments joined with fasteners to form a ring. The cameras are placed in a pentagonal formation around the ring, with gaps for the lateral handles of the surgical light (see Fig.~\ref{fig:overview} top right). Each camera is mounted via an adjustable locking hinge, positioning its field of view at an approximately 30$^{\circ}$ inclination from the horizontal so that the lines of sight converge roughly one meter below the light. This arrangement maximizes the likelihood of overlapping camera views, reducing the chance that a single occlusion obscures all viewpoints. All additional electronics are fastened to the fixture's upper surface. Three toggle clamps secure it to the surgical lamp, allowing the entire assembly to be attached or detached via quick-release levers. The supplementary document provides the bill of materials (Supplementary Figure~S1) and the assembly and component drawings (Supplementary Figures~S2--S25).

\subsubsection{Surgical Pointer and Static Reference}

To estimate the pose of the surgical instrument relative to the reference fixed to the patient, planar fiducial markers of the ArUco type \citep{Garrido2014} are employed, which are detected using OpenCV \citep{opencv_library}. To increase resilience against marker-level occlusion and rotational ambiguity, multiple markers are attached to a regular dodecahedron, as demonstrated by successful approaches in the literature \citep{Wu2017, Garcia2023}. An ArUco marker is centered on each face of the dodecahedron, except for the base, which attaches to the instrument handle or the reference shaft (see Fig.~\ref{fig:overview} bottom left). The geometry of the dodecahedron ensures that at least three faces remain visible from most viewing angles, enabling more robust pose estimation compared to simpler Platonic solids.

A 3D-printed part is fabricated to integrate the dodecahedron with a handle featuring a threaded insert for attachment of custom surgical instruments. A metal tip similar to pointer instruments employed in surgical navigation is affixed at the handle’s distal end, enabling precise sampling of measurement points and trajectories. The dodecahedron has a 77 mm circumscribed diameter to accommodate the placement of 24$\times$24~mm ArUco markers on each face, balancing reliable detection with a user-friendly form factor. The surgical pointer features a handle diameter of 5.4 mm and a total length of 220 mm from the center of the dodecahedron's top face to the end of the metal tip.

A similar dodecahedron of 112 mm circumscribed diameter is used for the static reference, affixed with 33$\times$33~mm ArUco markers. In our experimental setup, the reference dodecahedron is rigidly connected to the landmarks being measured. In an actual clinical workflow, the reference dodecahedron is fixed to a head clamp or an equivalent patient immobilizing structure, enabling tracking of the pointer relative to the patient irrespective of camera movement.

\subsubsection{Electronics and Optics}

In selecting the electronics and optics, the objective is to balance image quality, field of view, and communication speed with modularity and affordability. Raspberry Pi single-board computers (1 GB) are selected for their low cost and ability to enable low-latency transmission over Ethernet, and are paired with MIPI camera modules. The MIPI interface supports direct camera integration with the selected Raspberry Pi boards. The Raspberry\,Pi Global Shutter Camera (\(5.02\,\mathrm{mm}\) sensor width, \(1456\times1088\) at 60\,fps, \(3.45\,\mu\mathrm{m}\) pixel size) is chosen for its official support and C/CS lens mount, allowing for focal length customization \citep{raspberryPiShutter}. A 6\,mm lens is selected to provide a wide field of view covering the head and the 220\,mm pointer at a 1\,m working distance. Focal length \(f\) is estimated by
\(
f=\frac{H\times\mathrm{WD}}{\mathrm{Width}}
\),
where \(H\) is the sensor width, \(\mathrm{WD}\) is the working distance, and \(\mathrm{Width}\) is the subject size. For the stated sensor width, focal length, and working distance, this approximation gives a horizontal field width of approximately 837\,mm.

\subsection{Calibration Pipeline}
\subsubsection{Multi-Camera Calibration}\label{sec:camera_calib}

Every camera $c_i$ defines a right-handed Cartesian frame
$\mathcal F_{c_i}$ whose origin is its optical centre,
$z$-axis points forward along the viewing direction, and
$x$- and $y$-axes coincide with the image axes
($+x$ to the right, $+y$ downwards) (see Fig.~\ref{fig:instrument_calibration}).
Throughout this work, the frame of camera~1,
$\mathcal F_{c_1}$, serves as the global coordinate frame: all extrinsic
transforms $\mbf E_i$ map the global frame $\mathcal F_{c_1}$ into each
camera frame $\mathcal F_{c_i}$, with $\mbf E_1=\mbf I_{4\times4}$,
and every 3D point written without an explicit
camera coordinate frame subscript is understood to be expressed in $\mathcal F_{c_1}$.

To calibrate the system we acquire images of a ChArUco board
(see Fig.~\ref{fig:instrument_calibration}) at many poses by positioning the
board throughout the system's field of view. The board itself carries a fixed coordinate frame
$\mathcal F_{\!b}$ whose origin is at its top left corner and whose $z$-axis is normal to the printed surface.
The 2D pixel position of the $k$-th ChArUco corner observed in
camera $i$ at video frame (time step) $t$ is denoted
$\mbf x_{k,t,i}^{\mathbf b}\!\in\!\mathbb R^{2}$, and the known 3D
location of the same corner in the board coordinate frame is
$\mbf X_{k}^{\mathbf b}\!\in\!\mathbb R^{3}$.

For each frame $t$, $\mbf C_t^{bc_i}$ and $\mbf r_t^{bc_i}$ describe the
calibration board's rotation and translation into camera $i$'s coordinate frame.
Intrinsic calibration jointly estimates the intrinsics $\mbf M_i$ of each
camera and these per-frame board poses by minimising the reprojection error between every detected
$\mbf{x}_{k,t,i}^{\mathbf b}$ and its pinhole projection
$\pi\!\bigl(\mbf C_t^{bc_i}\mbf X_{k}^{\mathbf b}+\mbf r_t^{bc_i};\mbf M_i\bigr)$.
Thus, the error is always measured between a 2D corner observed in the
image and the same corner re-projected with the current camera intrinsics.
In our notation, the symbol
\[
  \mbf M_i \;:=\;
  \bigl(\,\tilde{\mbf M}_i,\;\boldsymbol d_i\bigr)
\]
always subsumes both the $3\times3$ pinhole camera matrix  
\[
  \tilde{\mbf M}_i \;=\;
  \begin{bmatrix}
    f_x & s   & c_x\\[2pt]
    0   & f_y & c_y\\[2pt]
    0   & 0   & 1
  \end{bmatrix},
\]
and the radial–tangential distortion vector  
\[
  \boldsymbol d_i \;=\; [\,k_1,k_2,k_3,p_1,p_2\,]^{\top}.
\]
Once $\tilde{\mbf M}_i$ and $\boldsymbol d_i$ have been identified, they enable accurate tracking through geometric projection and distortion correction.

Extrinsic calibration follows in two stages.
First, for each camera $i\neq1$, stereo calibration uses board corners
simultaneously visible in cameras $1$ and $i$. With both cameras' intrinsics
$\mbf M_1$ and $\mbf M_i$ fixed, the camera-to-camera transform
$\mbf E_{1i}$ and the board pose in each shared frame are jointly estimated
by minimising reprojection error in both cameras.
A subsequent global bundle-adjustment jointly refines the intrinsics $\mbf M_i$
and extrinsics $\mbf E_i$, with $\mbf E_1=\mbf I_{4\times4}$ fixed.
At each objective evaluation, board corners are triangulated in the global
frame (see Alg.~\ref{alg:stereo_tracking}) and board poses are estimated
using OpenCV's Perspective-n-Point (PnP) implementation~\citep{opencv_library},
both using the current calibration parameters. Each triangulated corner is
transformed into camera $i$'s frame using $\mbf E_i$ before comparison with
the corresponding corner predicted by that camera's PnP board pose.
The sum of squared 3D discrepancies across observed corners, frames, and
cameras is minimised, as summarised in Algorithm~\ref{alg:camera_calibration}.

\begin{algorithm}[!t]
\caption{Multi-Camera Calibration}
\label{alg:camera_calibration}
\begin{algorithmic}
\Statex \textbf{Input:} Synchronized frames $t\in\{1,\dots,T\}$ from cameras $i\in\{1,\dots,N\}$; known 3D ChArUco board corners $\mathbf{X}_{k}^{\mathrm b}$.
\Statex \textbf{Output:} Camera intrinsics $\mathbf{M}_i$ and extrinsics $\mathbf{E}_i$ mapping global coordinates into each camera's frame.

\Statex \textbf{1.\;Pre-processing}
\State Detect 2D corners $\mathbf{x}_{k,t,i}^{\mathrm b}$ in every image.
\State Keep frames where the board is observed by at least two cameras.

\Statex \textbf{2.\;Intrinsic calibration}
\For{$i=1$ to $N$}
  \State Jointly estimate $\mathbf{M}_i$ and the per-frame board poses
  $\{\mathbf{C}_t^{bc_i},\mathbf{r}_t^{bc_i}\}_t$ by minimising
  \Statex \hspace{\algorithmicindent}$\displaystyle \sum_{k,t}
  \Bigl\|
    \mathbf{x}_{k,t,i}^{\mathrm b} -
    \pi\!\bigl(\mathbf{C}_t^{bc_i}\mathbf{X}_{k}^{\mathrm b}
      +\mathbf{r}_t^{bc_i};\,\mathbf{M}_i\bigr)
  \Bigr\|^{2}$
\EndFor

\Statex \textbf{3.\;Pair-wise extrinsic initialization}
\State $\mathbf{E}_{1} \gets \mathbf{I}_{4\times4}$ \Comment{reference camera pose}
\For{$i=2$ to $N$}
  \State With $\mathbf{M}_1$ and $\mathbf{M}_i$ fixed, jointly estimate
  $\mathbf{E}_{1i}$ and the board pose in each shared frame by minimising
  reprojection error in both cameras.
  \State $\mathbf{E}_{i} \gets \mathbf{E}_{1i}$
\EndFor

\Statex \textbf{4.\;Global bundle adjustment}
\State With $\mathbf{E}_1=\mathbf{I}_{4\times4}$ fixed, jointly refine
$\{\mathbf{M}_i,\mathbf{E}_i\}$ by minimising
\Statex \hspace{\algorithmicindent}$\displaystyle \sum_{k,t,i}
  \Bigl\|
    \mathbf{C}^{c_1c_i}\tilde{\mathbf{X}}_{k,t}^{\mathrm b}
    +\mathbf{r}^{c_1c_i} -
    \bigl(\mathbf{C}_{t}^{bc_i}\mathbf{X}_{k}^{\mathrm b}
      +\mathbf{r}_{t}^{bc_i}\bigr)
  \Bigr\|^{2}$
\Statex where $(\mathbf{C}^{c_1c_i},\mathbf{r}^{c_1c_i})$ are the rotation
and translation of $\mathbf{E}_i$, $\tilde{\mathbf{X}}_{k,t}^{\mathrm b}$ is
the triangulated corner in the global frame, and
$(\mathbf{C}_{t}^{bc_i},\mathbf{r}_{t}^{bc_i})$ is the board pose estimated
by PnP for camera~$i$. Triangulation and PnP are recomputed using the current
calibration parameters at each objective evaluation; the sum includes only
observed corners.

\State \Return $\{\mathbf{M}_i,\mathbf{E}_i\}$
\end{algorithmic}
\end{algorithm}

\subsubsection{Instrument Calibration}\label{sec:instrument_calib}

Accurate tracking requires a good estimate of the relative 3D positions of every ArUco marker corner on the pointer and static reference: inaccurate estimates translate directly into poor pose estimation. We therefore take video of the tags and use bundle adjustment to arrive at a least-squares estimate that does not depend on precise manufacturing. This approach refines the initial corner coordinates on both instruments.

The instrument coordinate frames are defined as follows: the origin of the pointer coordinate frame $\mathcal F_{p}$ is at the physical pointer tip; the origin of the reference coordinate frame $\mathcal F_{r}$ is at the centre of its uppermost marker. Let $\mbf X_{k}^{\mathbf d,\mathrm{init}}\!\in\!\mathbb R^{3}$ be the initial coordinate of dodecahedron marker corner $k$ in its instrument frame ($\mathcal F_{p}$ or $\mathcal F_{r}$).

\begin{figure}[pos=!tbp]
  \centering
  \includegraphics[width=\linewidth]{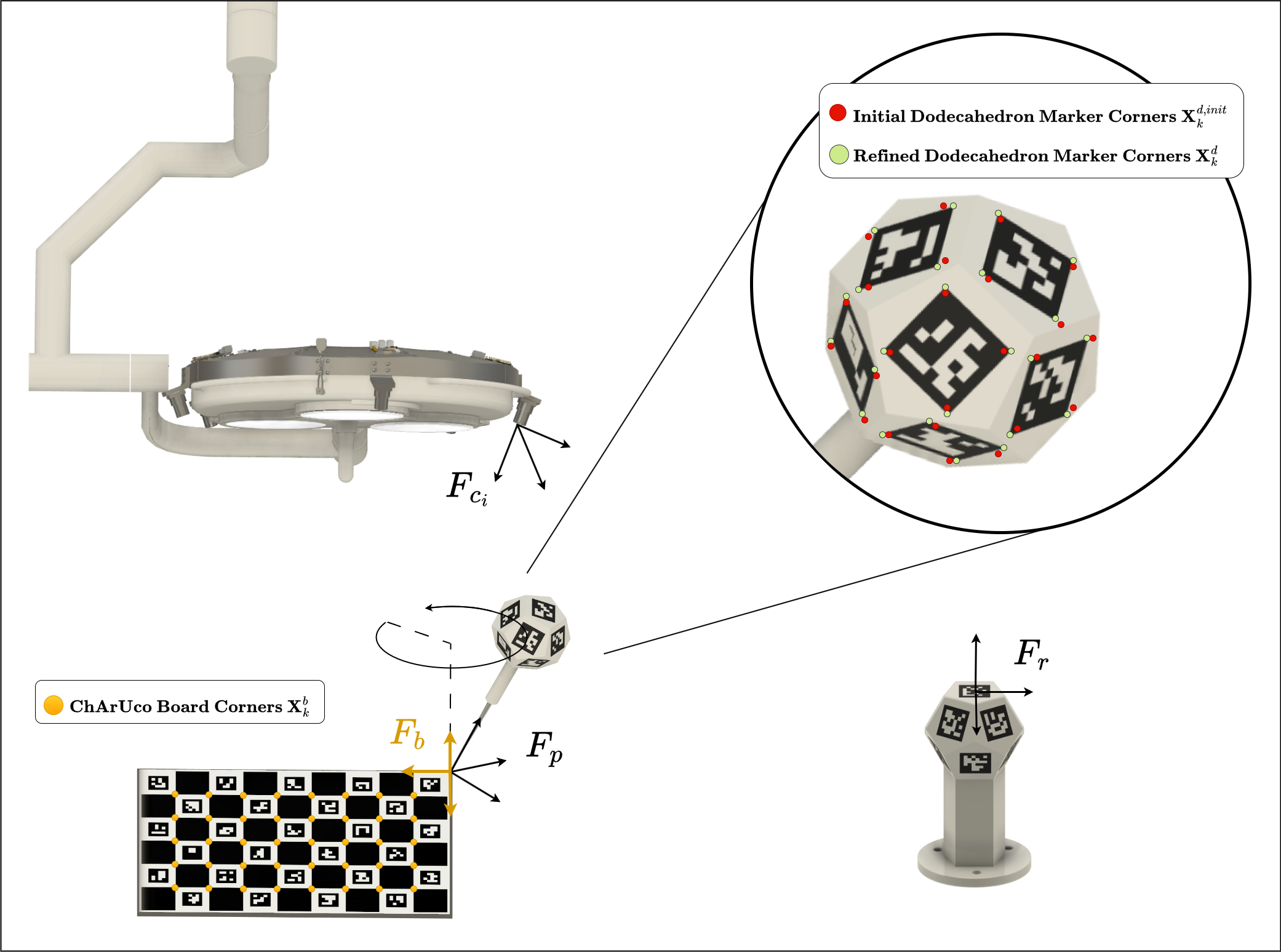}
  \caption{Instrument calibration schematic. The pointer is pivoted and rotated about its axis on the
  ChArUco board origin (bottom left) while video frames are recorded.  After initialisation the dodecahedron marker corner coordinates are refined.}
  \label{fig:instrument_calibration}
\end{figure}

During pointer calibration the
tip is placed on the origin of the fixed board frame $\mathcal F_{b}$
(see Fig.\,\ref{fig:instrument_calibration}) and the shaft is pivoted and rotated about its axis while
$T$ synchronised video frames are recorded from each camera. The initial tip position $\mbf p^{\mathrm{init}}$ is estimated by averaging the board origins detected across all $T$ frames.

For every frame $t$ and camera $i$ the pixel coordinates
$\mbf x_{k,t,i}^{\mathbf d}$ of each visible corner are detected and,
using the previously calibrated intrinsics $\mbf M_i$ and extrinsics
$\mbf E_i$, are undistorted and triangulated (see Alg.~\ref{alg:stereo_tracking}) to the global points
$\tilde{\mbf X}_{k,t}^{\mathbf d}$. The initial corner set and the measured cloud are aligned independently
in every video frame. For the pointer corner set $\mathcal{P}$ we solve for a rotation
$\mbf U_t\!\in\!SO(3)$ about the fixed tip while for the reference corner set $\mathcal{R}$ we solve
for a full rigid transformation $(\mbf U_t,\mbf v_t)\!\in\!SE(3)$. Here $SO(3)=\{\mbf U\!\in\!\mathbb R^{3\times3}\mid
             \mbf U^{\trans}\mbf U=\mbf I,\,
             \det\mbf U=1\}$ is the set of all 3D rotation matrices,
and $SE(3)=\{(\mbf U,\mbf v)\mid \mbf U\!\in\!SO(3),\;\mbf v\!\in\!\mathbb R^{3}\}$
is the set of all rigid-body motions composed of a rotation $\mbf U$ and
a translation $\mbf v$.

The relative corner coordinates are refined across all video frames by minimising

\[
\begin{aligned}
\arg\min_{\mbf X,\,\mbf p}
\sum_{k,t\in\mathcal P}
  \bigl\|
    \mbf U_t\,\mbf X_{k}^{\mathbf d}
    +\mbf p
    -\tilde{\mbf X}_{k,t}^{\mathbf d}
  \bigr\|^{2},
&&\text{for the pointer},
\\[6pt]
\arg\min_{\mbf X}
\sum_{k,t\in\mathcal R}
  \bigl\|
    \mbf U_t\,\mbf X_{k}^{\mathbf d}
    +\mbf v_t
    -\tilde{\mbf X}_{k,t}^{\mathbf d}
  \bigr\|^{2},
&&\text{for the reference}.
\end{aligned}
\]
Instrument calibration outputs the refined corner set
$\{\mbf X_{k}^{\mathbf d}\}$ in its respective instrument
frame ($\mathcal F_{p}$ or $\mathcal F_{r}$).

\begin{algorithm}[!t]
\caption{Instrument Calibration}
\label{alg:instrument_calibration}
\begin{algorithmic}
\Statex \textbf{Input:} Synchronized video frames $t\in\{1,\dots,T\}$ from cameras $i\in\{1,\dots,N\}$;
ChArUco board corners $\mbf X_k^{\mathbf b}$ (for pointer calibration only);
camera intrinsics $\mbf M_i$ and extrinsics $\mbf E_i$ (Alg.~\ref{alg:camera_calibration});
initial dodecahedron marker corners $\mbf X_{k}^{\mathbf d,\mathrm{init}}$.
\Statex \textbf{Output:} Refined corners $\mbf X_{k}^{\mathbf d}$.

\Statex \textbf{1.\;Pre-processing}
\If{pointer}
  \State Initialize $\mbf p^{\mathrm{init}}$ as the mean ChArUco-board origin across
  \State all frames
\EndIf

\Statex \textbf{2.\;Marker detection and triangulation}
\For{$t=1$ to $T$}
  \State Detect $\mbf x_{k,t,i}^{\mathbf d}$ in each camera $i$; keep corners seen by $\ge 2$
  \State cameras
  \ForAll{visible corners $k$}
    \State $\tilde{\mbf X}_{k,t}^{\mathbf d} \gets
      \operatorname{triangulate}\!\bigl(\mbf M_i,\mbf E_i,\mbf x_{k,t,i}^{\mathbf d}\bigr)$
  \EndFor
\EndFor

\Statex \textbf{3.\;Frame-wise alignment}
\For{$t=1$ to $T$}
  \If{pointer}
    \State Estimate $\mbf U_t$ aligning
      $\{\mbf X_{k}^{\mathbf d}\}$ to $\{\tilde{\mbf X}_{k,t}^{\mathbf d}-\mbf p^{\mathrm{init}}\}$
    \State $\Delta\mbf X^{\mathbf d}_{k,t} \gets \mbf U_t^{\top}
      \bigl(\mbf U_t\mbf X_{k}^{\mathbf d}
      +\mbf p^{\mathrm{init}}-\tilde{\mbf X}_{k,t}^{\mathbf d}\bigr)$
  \Else
    \State Estimate $(\mbf U_t,\mbf v_t)$ aligning the two sets
    \State $\Delta\mbf X^{\mathbf d}_{k,t} \gets \mbf U_t^{\top}
      \bigl(\mbf U_t\mbf X_{k}^{\mathbf d}
      +\mbf v_t-\tilde{\mbf X}_{k,t}^{\mathbf d}\bigr)$
  \EndIf
\EndFor

\State $\mbf X_{k}^{\mathbf d} \gets \mbf X_{k}^{\mathbf d}
       - \dfrac{\sum_{t\in\mathcal T_k}w_{k,t}\Delta\mbf X^{\mathbf d}_{k,t}}
                {\sum_{t\in\mathcal T_k}w_{k,t}}$
\Statex where $\mathcal T_k$ contains the frames in which corner $k$ is
triangulated, and $w_{k,t}=\binom{n_{k,t}}{2}$, with $n_{k,t}$ the number
of cameras observing its marker in frame $t$. The errors
$\Delta\mbf X^{\mathbf d}_{k,t}$ are expressed in the instrument frame.
\State Iterate to convergence.
\State Reject a marker observation in a frame if any component of its
corner errors exceeds $2\sigma$ in magnitude, where $\sigma$ is the
standard deviation across all valid corner-error components.

\Statex \textbf{4.\;Bundle adjustment}
\If{pointer}
  \State $(\{\mbf X_{k}^{\mathbf d}\},\mbf p) \gets
  \arg\min_{\{\mbf X_{k}^{\mathbf d}\},\,\mbf p}$
  \Statex \hspace{\algorithmicindent}$\displaystyle
  \sum_{k,t}\bigl\|\mbf U_t\mbf X_{k}^{\mathbf d}
                 + \mbf p
                 - \tilde{\mbf X}_{k,t}^{\mathbf d}\bigr\|^{2}$
\Else
  \State $\mbf X_{k}^{\mathbf d} \gets
  \arg\min_{\{\mbf X_{k}^{\mathbf d}\}}
  \sum_{k,t}\bigl\|\mbf U_t\mbf X_{k}^{\mathbf d}
                 + \mbf v_t
                 - \tilde{\mbf X}_{k,t}^{\mathbf d}\bigr\|^{2}$
\EndIf
\Statex Recompute frame alignments at each objective evaluation.

\State \Return $\{\mbf X_{k}^{\mathbf d}\}$
\end{algorithmic}
\end{algorithm}

\subsection{Pose Estimation Methods}

\subsubsection{Multi-View Monocular Pose Fusion}\label{sec:mono_tracking}

Multi-view monocular pose fusion estimates the pointer pose relative to the static reference by computing independent transformations from each camera view and combining them with an adaptive Kalman filter. Let
$\mbf r^{pr}\!\in\!\mathbb R^{3}$ be the position of $\mathcal F_{p}$
with respect to $\mathcal F_{r}$ and
$\mbf C^{pr}\!\in\!SO(3)$ the corresponding orientation.  
Together they form the homogeneous pose

\begin{equation}\label{eq:pose_se3}
  \mbf T^{pr}=%
  \begin{bmatrix}
    \mbf C^{pr} & \mbf r^{pr}\\
    \mathbf 0^{\trans} & 1
  \end{bmatrix}\in SE(3).
\end{equation}

Here, $\mbf T^{ab}$ maps homogeneous coordinates from frame
$\mathcal F_a$ to frame $\mathcal F_b$.

For every camera $c_i$ with an unobstructed view, the ArUco detector \citep{Garrido2014}
yields 2D marker corners; a PnP algorithm returns the
monocular pointer and reference poses
$\mbf T^{pc_i}$ and $\mbf T^{rc_i}$ in $\mathcal F_{c_i}$.
Their composition gives an independent estimate of the
pointer–to–reference transform

\begin{equation}\label{eq:rel_ori}
  \mbf T^{pr}_{i} \;=\;
  \bigl(\mbf T^{rc_i}\bigr)^{-1}\,
  \mbf T^{pc_i}.
\end{equation}

The set of computed transforms
$\{\mbf T^{pr}_{i}\}$ is fused frame-by-frame with an
adaptive Kalman filter, yielding the final smoothed estimate
${\mbf T}^{pr}_{final}$ delivered to the navigation system for each frame.

\begin{figure}[pos=!tbp]
\centering
\includegraphics[width=\linewidth]{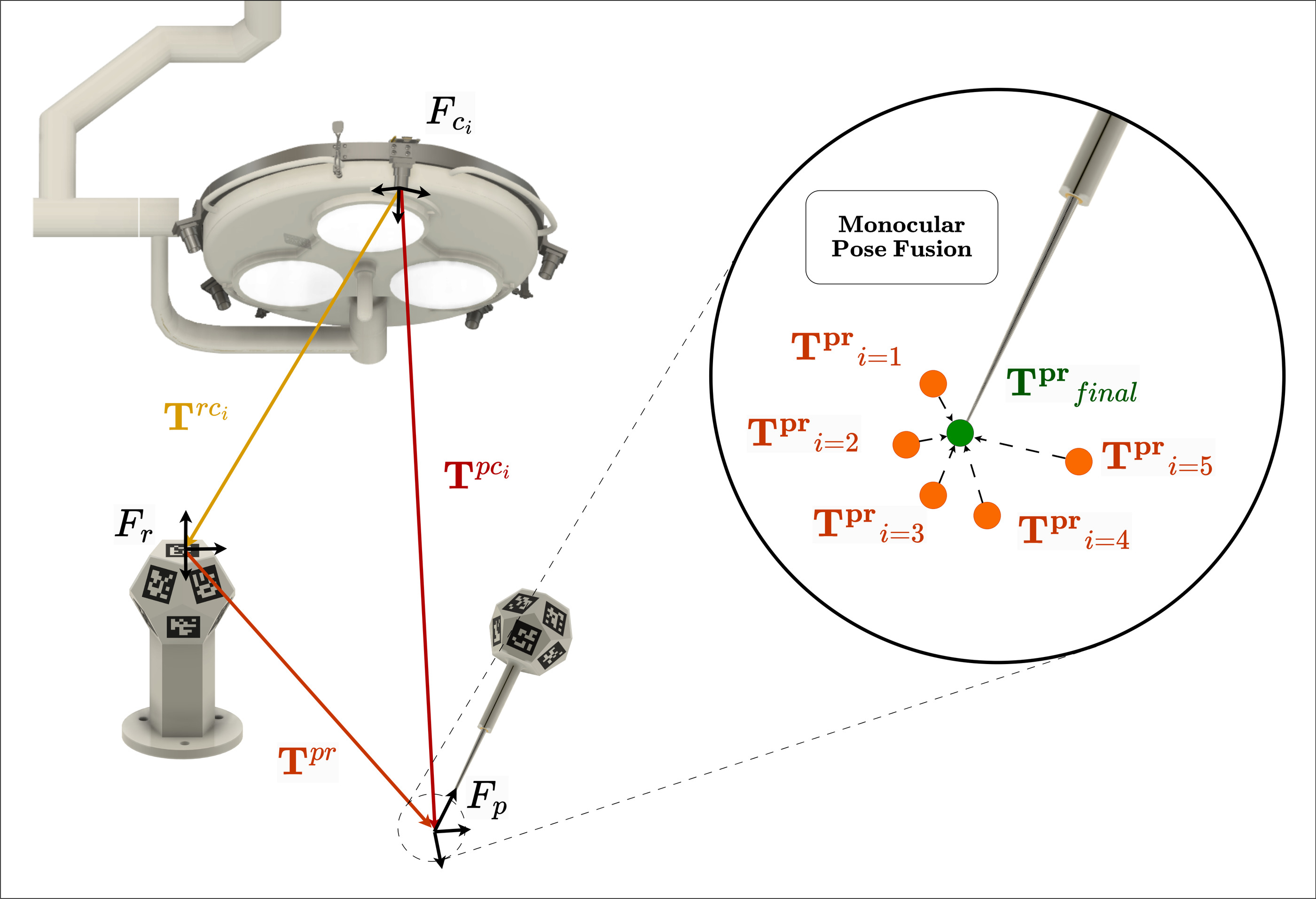}
\caption{Multi-view monocular pose fusion schematic. For each camera that views the surgical pointer and static reference, an independent monocular pose measurement is computed through the PnP algorithm. At each frame, an adaptive Kalman filter is used to fuse all available measurements.}
\label{fig:mono_tracking}
\end{figure}

The adaptive Kalman filter is used to improve pose estimation accuracy under measurement noise. While \citet{Popescu2020} use a zero velocity model, our approach updates both position and velocity in the state vector. Unlike their method, which updates noise covariances over all past data, we dynamically adjust the pose block of the process noise covariance matrix $\mbf{Q}_t$ and the measurement noise covariance matrix $\mbf{R}_t$ at each frame $t$ using valid pose estimates from a rolling window of up to 30 frames. This allows the filter to respond more effectively to recent changes in noise caused from occlusions, pointer motion, and variations in lighting conditions.

A process model is defined for the surgical pointer. Due to the steady motion of the pointer, especially around points of interest, a constant velocity model is assumed. Moreover, as no proprioceptive measurement of the pointer itself is available, any estimates of the acceleration are inherently noisy. Thus, the state is defined to only consist of the translational and rotational position and velocity of the pointer. Specifically, the state vector $\mbf{x}_t$ at frame $t$ is represented as,

\begin{equation} 
\mbf{x}_t = \begin{bmatrix} (\mbf{r}^{pr})^{\trans} & (\mbs{\theta}^{pr})^{\trans} & (\dot{\mbf{r}}^{pr})^{\trans} & (\dot{\mbs{\theta}}^{pr})^{\trans} \end{bmatrix}^\trans ,
\end{equation} where $\mbf{r}^{pr}$ represents the position of the pointer relative to the reference as defined in Eq.~\ref{eq:pose_se3}, $\mbs{\theta}^{pr} = (\chi^{pr}, \phi^{pr}, \psi^{pr})$ is the orientation of the pointer relative to the reference's coordinate frame expressed as Euler angles, and the dot notation denotes the corresponding velocities. For simplicity, all coordinate frame subscripts of the position and orientation variables are omitted with the understanding that they are expressed relative to the static reference's coordinate frame, unless otherwise stated.

The state transition model is written in discrete time as, 

\begin{equation} \mbf{x}_t = \mbf{A} \mbf{x}_{t-1} + \mbf{w}_{t-1}, \quad \quad \mbf{w}_{t-1} \sim \mathcal{N}(0, \mbf{Q}_{t-1}), \end{equation} where $\mbf{A}$ is the state transition matrix, and $\mbf{w}_{t-1}$ is the process noise, assumed to be zero-mean, white Gaussian noise with covariance matrix $\mbf{Q}_{t-1}$. The state transition matrix for a constant velocity model is

\begin{equation}
\mbf{A} = \begin{bmatrix} 
\mbf{I}_6 & \mbf{I}_6 \Delta t\\
\mbf{0}_6 & \mbf{I}_6
\end{bmatrix},
\end{equation} where $\mbf{I}_6$ is the $6 \times 6$ identity matrix, $\mbf{0}_6$ is the $6 \times 6$ zero matrix, and $\Delta t$ is the time interval between measurements.

At each frame $t$, measurements are obtained from multiple synchronised cameras through an electrical signal. Each camera that detects both the pointer and the static reference is considered active and provides an independent pose estimate of the pointer relative to the reference using the PnP algorithm. The measurement vector $\mbf{y}_t$ consists of the observed positions and orientations from all active cameras. The measurement vector from the $i$-th active camera is written as

\begin{equation}
\mbf{y}_{t,i}=
\begin{bmatrix}
\mbf{r}_{t,i}^{\,pr}\\
\boldsymbol{\theta}_{t,i}^{\,pr}
\end{bmatrix},
\qquad
i=1,\dots,n .
\end{equation}

The overall measurement vector $\mbf{y}_t$ is constructed by stacking the measurements from all $n$ active cameras such that

\begin{equation}
\mbf{y}_{t}=
\begin{bmatrix}
\mbf{y}_{t,1}\\
\vdots\\
\mbf{y}_{t,n}
\end{bmatrix}.
\end{equation}

The measurement model relates the state vector to the measurements through the measurement matrix $\mbf{H}_t$ such that

\begin{equation} \mbf{y}_{t}= \mbf{H}_{t}\,\mbf{x}_{t}+ \mbf{n}_{t},
\quad \quad
\mbf{n}_{t}\!\sim\!\mathcal{N}(\mathbf 0,\mbf{R}_{t}), \end{equation} where $\mbf{n}_t$ is the measurement noise, assumed to be zero-mean Gaussian with covariance $\mbf{R}_t$.

Because the position and orientation are measured but their corresponding velocities are not, the measurement matrix $\mbf{H}_t$ at video frame $t$ is constructed by vertically stacking submatrices corresponding to each camera $i$ as

\begin{equation}
\mbf{H}_{t}=
\begin{bmatrix}
\mbf{I}_{6} & \mbf{0}_{6}\\
\vdots & \vdots\\
\mbf{I}_{6} & \mbf{0}_{6}
\end{bmatrix}.
\end{equation} The number of rows in $\mbf{H}_t$ is $6n$, where $n$ is the number of active cameras at frame $t$.

To adapt the filter to observed pose variability, the pose block of the process noise covariance matrix $\mbf{Q}_t$ and the measurement noise covariance matrix $\mbf{R}_t$ are updated at each video frame.
For each camera, pose variability is estimated from valid measurements
within a rolling window of up to 30 frames. Missing poses are omitted,
and the available history is used during startup.
Let $\mathcal W_{t,i}$ contain the frame indices of the valid poses in
this window for active camera $i$, and let $m_{t,i}=|\mathcal W_{t,i}|$.
The measurement noise covariance $\mbf R_{t,i}$ is diagonal, with a separate
variance for each of the six pose components:

\begin{equation}
\begin{aligned}
\mbf R_{t,i} &= \operatorname{diag}\!\bigl(s_{t,i,1}^{2},\dots,s_{t,i,6}^{2}\bigr),\\
s_{t,i,j}^{2} &= \frac{1}{m_{t,i}}\sum_{\tau\in\mathcal W_{t,i}}
  \bigl(y_{\tau,i,j}-\bar y_{t,i,j}\bigr)^{2}.
\end{aligned}
\end{equation}
Here $y_{\tau,i,j}$ is component $j$ of $\mbf y_{\tau,i}$ and
$\bar y_{t,i,j}$ is its mean over $\mathcal W_{t,i}$. These variances reflect both measurement fluctuations and actual tool motion and are used as empirical covariance estimates.

The overall measurement noise covariance matrix \( \mbf{R}_t \) is then constructed by placing the individual \( \mbf{R}_{t,i} \) matrices along the diagonal such that

\begin{equation}
\mbf{R}_{t} =
\operatorname{diag}\!\bigl(
  \mbf{R}_{t,1},\;
  \mbf{R}_{t,2},\;
  \dots,\;
  \mbf{R}_{t,n}
\bigr),
\end{equation} where $n$ is the number of active cameras providing measurements at frame $t$.

The filter uses camera-derived poses without auxiliary inertial
measurements. For this camera-only implementation, we adopt an empirical
covariance rule: the adaptive pose-noise block is set to the mean of the
active cameras' pose-variability matrices, while the velocity-noise block
remains fixed. Averaging sets prediction uncertainty on the scale of
observed pose variability without increasing that scale solely with
camera count:
\begin{equation}
\bar{\mbf{R}}_{t}= \tfrac{1}{n}\sum_{i=1}^{n}\mbf{R}_{t,i}, 
\qquad 
\mbf{Q}_{t}= 
\begin{bmatrix} 
\bar{\mbf{R}}_{t} & \mbf{0}_{6}\\ 
\mbf{0}_{6} & 0.001\,\mbf{I}_{6}
\end{bmatrix}.
\end{equation}
The velocity estimates and state covariance are initialized to zero.
A fixed nonzero velocity process-noise term is included to allow pose
measurements to update the velocity estimates. Each of the six velocity
process-noise variances is set to 0.001 per prediction step in both the
adaptive and non-adaptive filters. The translational and angular velocity
variances are expressed in $(\mathrm{mm}/\mathrm{s})^{2}$ and
$(\mathrm{deg}/\mathrm{s})^{2}$, respectively.

The standard Kalman filter equations \citep{Kalman1960} are then applied to estimate the state $\mbf{x}_t$ at each frame. The prediction step is defined as
 
\begin{align}\label{eq:state_prediction}
  \check{\mbf{x}}_{t} &= \mbf{A}\,\hat{\mbf{x}}_{t-1},\\
  \check{\mbf{P}}_{t} &= \mbf{A}\,\hat{\mbf{P}}_{t-1}\,\mbf{A}^{\!\top}
                        + \mbf{Q}_{\,t-1},
\end{align} where $\check{\mbf{x}}_{t}$ is the predicted state estimate, and $\check{\mbf{P}}_{t}$ is the predicted estimate covariance.

The update step is

\begin{align}\label{eq:state_update}
  \mbf{K}_{t} &=
    \check{\mbf{P}}_{t}\,\mbf{H}_{t}^{\!\top}
    \bigl(\mbf{H}_{t}\check{\mbf{P}}_{t}\mbf{H}_{t}^{\!\top}
          +\mbf{R}_{t}\bigr)^{-1},\\
  \hat{\mbf{x}}_{t} &=
    \check{\mbf{x}}_{t}
    + \mbf{K}_{t}\bigl(\mbf{y}_{t}-\mbf{H}_{t}\check{\mbf{x}}_{t}\bigr),\\
  \hat{\mbf{P}}_{t} &=
    \bigl(\mbf{I}-\mbf{K}_{t}\mbf{H}_{t}\bigr)\check{\mbf{P}}_{t},
\end{align} where $\mbf{K}_t$ is the Kalman gain, $\hat{\mbf{x}}_{t}$ is the updated state estimate, and $\hat{\mbf{P}}_{t}$ is the updated estimate covariance.

By updating the pose block of $\mbf{Q}_t$ and $\mbf{R}_t$ at each frame, the Kalman filter adapts its covariance weighting to the observed pose variability. We compare this adaptive approach with a constant-velocity Kalman filter using static covariance matrices. Noise profiling of stationary recordings provides fixed measurement noise covariance matrices $\bar{\mbf{R}}_{i}$ for each camera and the pose block of $\mbf{Q}$ for the non-adaptive filter, retaining the same fixed velocity-noise block. Details of this procedure are provided in Appendix~\ref{sec:noise_profiling}.

\subsubsection{Multi-View Triangulation}
Multi-view triangulation leverages the rigid positioning between cameras in the system to estimate the pose of the surgical pointer relative to the static reference. Marker corners on the surgical pointer and the static reference are first
detected using OpenCV's ArUco tools
\citep{opencv_library} and undistorted using the intrinsic parameters
described in Section~\ref{sec:camera_calib}.
For each corner $k$ that is observed by at least two cameras
(index-set $\mathcal I_k\subseteq\{1,\dots,N\}$), the 3D point
$\tilde{\mbf X}_{k}^{\mathbf d}$ in the global (camera 1) coordinate frame
is estimated by multi-view linear triangulation from the undistorted image
coordinates $\mbf x_{k,i}^{\mathbf d}$, $i\!\in\!\mathcal I_k$, and the
camera projection matrices, using singular value decomposition.
Each corner is assigned a registration weight
$w_k=\tfrac12\,|\mathcal I_k|\!\bigl(|\mathcal I_k|\!-\!1\bigr)$, equal to the
number of camera pairs that observe it.

The triangulated corners are reprojected into the undistorted image planes
of the cameras observing each marker. If the Euclidean reprojection error
of any corner exceeds 8 pixels in any of these cameras, all four corners
of that marker are discarded for that frame. Weighted point cloud
registration of the remaining corners is then performed to minimise
\[
\arg\min_{\mbf C\in SO(3),\,\mbf r\in\mathbb R^{3}}
\sum_k w_k\,
\bigl\|\mbf C\,\mbf X_k^{\mathbf d}+\mbf r-\tilde{\mbf X}_{k}^{\mathbf d}\bigr\|^{2}.
\]
to register the refined instrument model
(Section~\ref{sec:instrument_calib}) to the triangulated cloud,
yielding the pointer pose
\((\mbf C_t^{pc_1},\mbf r_t^{pc_1})\) and reference
pose \((\mbf C_t^{rc_1},\mbf r_t^{rc_1})\) with respect to camera 1.
The desired relative transform \(\mbf T^{pr}_t\) is obtained by composition, exactly as in Eq.~\eqref{eq:rel_ori} of
Section~\ref{sec:mono_tracking}.
The complete procedure is described in Algorithm~\ref{alg:stereo_tracking}.

\begin{algorithm}[!t]
\caption{Multi-View Triangulation (one frame)}
\label{alg:stereo_tracking}
\begin{algorithmic}
\Statex \textbf{Input:} Synchronized images from $i\in\{1,\dots,N\}$ cameras; camera intrinsics $\mbf M_i$ and extrinsics $\mbf E_i$ (Alg.~\ref{alg:camera_calibration}); dodecahedron marker corners $\mbf X_{k}^{\mathbf d}$ (for pointer \& reference; Alg.~\ref{alg:instrument_calibration}).
\Statex \textbf{Output:} Relative pose $\mbf T^{pr}$ of the pointer w.r.t.\ the reference.

\Statex \textbf{1.\;Corner detection}
\State Detect 2D image points $\mbf x_{k,i}^{\mathbf d}$ in every camera $i$; keep points seen by at least two cameras and undistort them

\Statex \textbf{2.\;Multi-view triangulation}
\ForAll{corner $k$}
  \State $\mathcal I_k \gets \{\, i \;|\; \mbf x_{k,i}^{\mathbf d}\ \text{exists} \,\}$
  \State $\tilde{\mbf X}_{k}^{\mathbf d} \gets
         \operatorname{triangulate}\bigl(\{\mbf M_i,\mbf E_i,\mbf x_{k,i}^{\mathbf d}\}_{i\in\mathcal I_k}\bigr)$
\EndFor

\Statex \textbf{3.\;Re-projection outlier rejection}
\ForAll{marker with four triangulated corners $\mathcal K$}
  \ForAll{$k\in\mathcal K$, $i\in\mathcal I_k$}
    \State $\Delta\mbf x_{k,i}^{\mathbf d} \gets
           \mbf x_{k,i}^{\mathbf d} - \pi\bigl(\tilde{\mbf X}_{k}^{\mathbf d};\,\tilde{\mbf M}_i,\mbf E_i\bigr)$
  \EndFor
  \If{any $\|\Delta\mbf x_{k,i}^{\mathbf d}\|_2>8$ pixels}
    \State Discard all four corners of this marker
  \EndIf
\EndFor
\Statex \textbf{4.\;Pose estimation}
\ForAll{corner $k$ not discarded}
  \State $w_k \gets \tfrac12\,|\mathcal I_k|\bigl(|\mathcal I_k|-1\bigr)$
\EndFor
\State $(\mbf C^{pc_1},\mbf r^{pc_1}) \gets
       \arg\min_{\{\mbf C,\mbf r\}}
       \sum_{k\in\mathcal P} w_k\,
       \|\mbf C\,\mbf X_{k}^{\mathbf d}+\mbf r-\tilde{\mbf X}_{k}^{\mathbf d}\|^{2}$
\State $(\mbf C^{rc_1},\mbf r^{rc_1}) \gets
       \arg\min_{\{\mbf C,\mbf r\}}
       \sum_{k\in\mathcal R} w_k\,
       \|\mbf C\,\mbf X_{k}^{\mathbf d}+\mbf r-\tilde{\mbf X}_{k}^{\mathbf d}\|^{2}$

\Statex \textbf{5.\;Relative pose composition}
\State $\mbf T^{pr} \gets (\mbf T^{rc_1})^{-1}\,\mbf T^{pc_1}$
\State where $\mbf T^{pc_1} = \begin{bmatrix}\mbf C^{pc_1} & \mbf r^{pc_1}\\ \mathbf 0^{\top} & 1\end{bmatrix}$ and
      $\mbf T^{rc_1} = \begin{bmatrix}\mbf C^{rc_1} & \mbf r^{rc_1}\\ \mathbf 0^{\top} & 1\end{bmatrix}$

\State \Return $\{\mbf T^{pr}\}$
\end{algorithmic}
\end{algorithm}

In the multi-view triangulation tracking method, the adaptive Kalman filter is applied to smooth the estimated pose trajectory. Unlike multi-view monocular pose fusion, where multiple pose estimates are merged and each camera’s measurement can be assigned an individual covariance, the filter here operates on a single triangulated pose estimate per frame. Consequently, the measurement noise covariance $\mathbf{R}_t$ is constructed from the recent history of this singular pose estimate and equals the $6\times6$ pose block of the process noise covariance $\mathbf{Q}_t$. The velocity-noise block remains fixed at $0.001\,\mathbf{I}_6$. Using the constant velocity dynamics described in Section~\ref{sec:mono_tracking}, the prediction step \eqref{eq:state_prediction} propagates the state forward in time, while the correction step \eqref{eq:state_update} refines the predicted pose with the measurement. This approach suppresses high-frequency noise and ensures that the tracking profile remains consistent with the expected motion dynamics, but it cannot selectively suppress cameras with poor views.

\section{Experimental Protocol}\label{sec:experiments}

\subsection{Experimental Setups}

\begin{figure}[pos=!tbp]
  \centering
  \includegraphics[width=\linewidth]{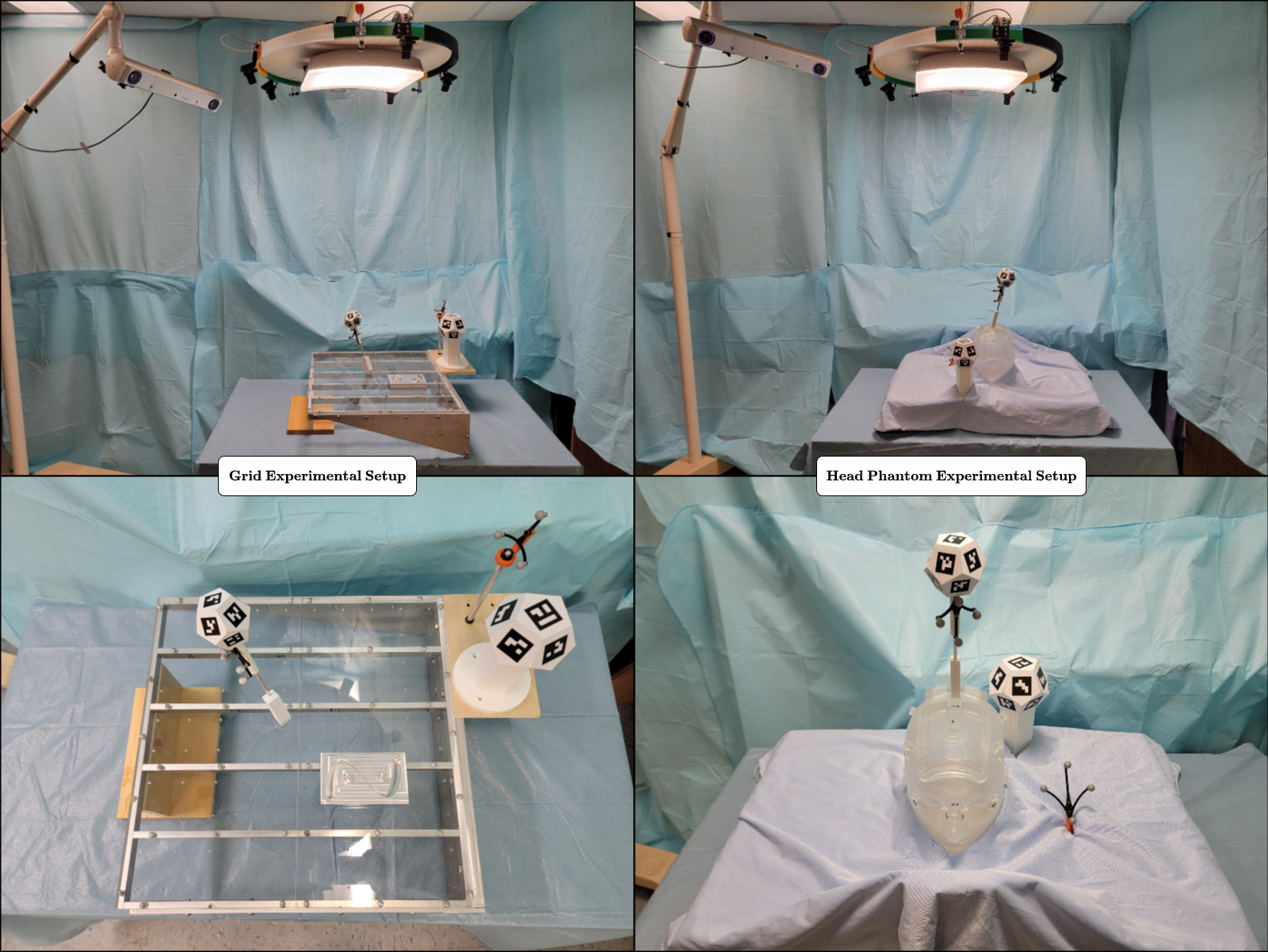}
  \caption{Experimental validation setups. Left column shows the experiment with the precision-machined grid. Right column shows the experiment with the SLA-printed head phantom. Both setups include the multi-camera fixture mounted on the surgical light stand-in, the surgical pointer, and the static reference. The fusionTrack 500 optical tracker (visible on the side) provides simultaneous measurements.}
  \label{fig:experimental_setups}
\end{figure}

The performance of the proposed system is evaluated through two experimental setups: a grid experiment designed to establish baseline accuracy and a head phantom experiment to better reflect the anatomical and operational conditions observed during IGNS procedures. A stand-in for a typical surgical light was manufactured to support the fixture and a light source. The stand-in measures 72 cm in diameter, derived from the specification sheet of our reference surgical light model \citep{infinium}. In both experiments, the ring fixture is mounted on the surgical light stand-in positioned 1 meter above the centre of the tracking volume. For benchmarking, a gold standard Atracsys fusionTrack 500 optical tracking system is placed adjacent to the phantom with a sensor-to-object distance of 1.4 meters per manufacturer specifications. Retroreflective sphere patterns are attached to the surgical pointer and next to the tracking volume, enabling simultaneous tracking by both systems. Figure~\ref{fig:experimental_setups} shows photos of both experimental setups with this consistent configuration.

\subsubsection{Grid Experiment}
The baseline accuracy of the proposed system is evaluated on a machined 400 mm × 400 mm grid \citep{Gerard2015} using landmark registration (see Appendix~\ref{sec:landmark_registration}). The grid features 25 uniformly spaced points, 9 of which are designated as fiducials. Each trial involves sampling the 9 fiducials to compute the fiducial registration error (FRE), and using the resulting transformation to calculate the TRE of the remaining 16 points. When collecting data from either a fiducial or target point, the tracked pointer is placed in a pointer holder that is inserted into a 2 mm diameter hole drilled at 100 mm center-to-center distance grid points using a numerically controlled milling machine, with random z-axis orientation at each point. This minimises noise due to movement and should yield the best accuracy estimate. To evaluate the TRE for dynamic trajectories, a block bearing an S-shaped curve with dimensions 100 mm x 50 mm x 15 mm is machined and placed at four positions along the surface of the grid as shown in Figure~\ref{fig:landmark}. Trajectory data was collected by holding the tracked pointer and sliding it along the S-shaped groove. This was designed to more realistically estimate tracking accuracy. Both multi-view monocular pose fusion and multi-view triangulation tracking modes are evaluated across 20 trials. The complete dataset comprises 5,000 synchronized multi-view videos across both experimental setups, all publicly available at \url{https://nist.mni.mcgill.ca/data/} for benchmarking alternative tracking algorithms.

\subsubsection{Head Phantom Experiment}
\begin{figure*}[width=\textwidth,pos=!tbp]
  \centering

  \begin{subfigure}[t]{\linewidth}
    \centering
    \includegraphics[width=\linewidth]{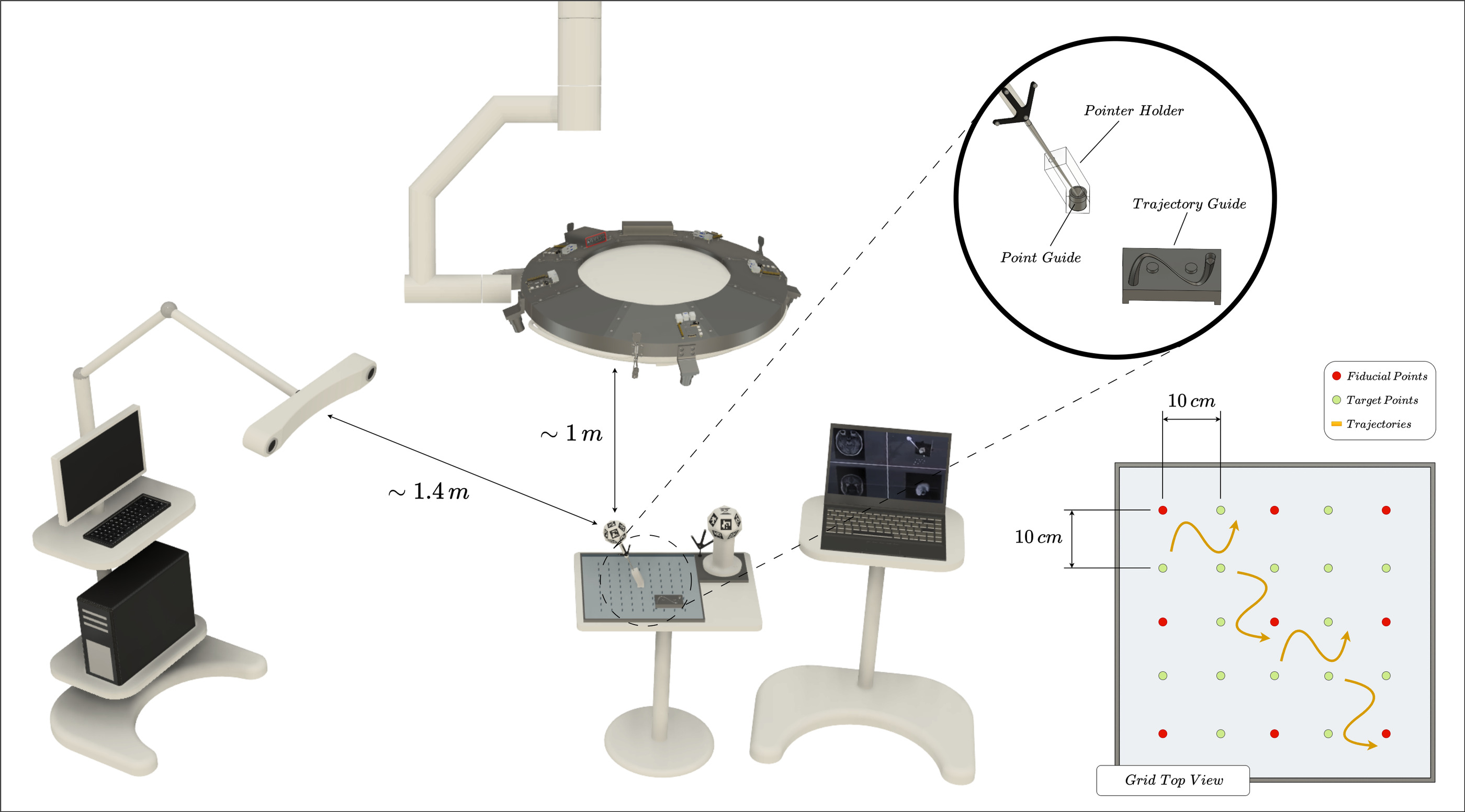}
    \caption{Machined grid used for landmark registration. It carries a set of 9 fiducial points (red), 16 target points (green), and 4 trajectories (orange). Each trial samples the fiducials to compute FRE and uses the transformation to calculate TRE on the remaining points. Trajectory data is collected along S-shaped grooves as illustrated.}
    \label{fig:landmark}
  \end{subfigure}

  \vspace{1em} %

  \begin{subfigure}[t]{\linewidth}
    \centering
    \includegraphics[width=\linewidth]{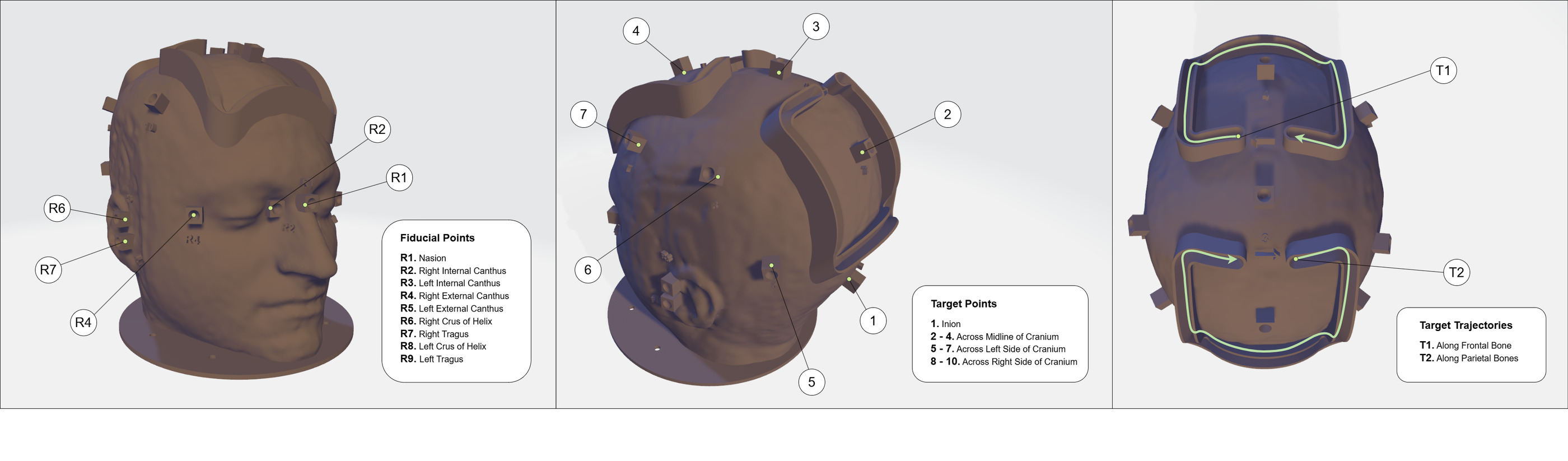}
    \caption{SLA-printed head phantom used for landmark registration. It carries 9 fiducial points (left), 10 target points (middle), and 2 trajectories (right). Each trial uses sampled fiducials to estimate a registration transform and compute FRE. The transform is then applied to the target points and trajectories to calculate their TRE.}
    \label{fig:head}
  \end{subfigure}

  \caption{Experimental setups for system evaluation. (a) Machined grid for baseline accuracy measurements. (b) SLA-printed head phantom for neurosurgery-representative validation.}
  \label{fig:experiment_setups}
\end{figure*}

To validate the system in conditions more representative of neurosurgical workflows, a complementary evaluation is performed using an SLA-printed head phantom (see Fig.~\ref{fig:head}). The phantom, derived from a 3T MRI scan, features 21 landmarks: 9 fiducial points corresponding to anatomical landmarks commonly used for fiducial registration in IGNS \citep{Taleb2023}, 10 target points distributed across the cranium, and two square grooves placed along the frontal and parietal bones for trajectory tracking. Each trial consists of sampling the 9 fiducials to calculate the FRE. The resulting transformation is used to compute the point TRE for the 10 target points and the trajectory TRE along the grooves. Results are aggregated over 20 trials, and detailed performance metrics for multi-view monocular pose fusion and multi-view triangulation are provided in Table~\ref{tab:results}. The model for the head phantom can be found at \url{https://nist.mni.mcgill.ca/software/}.

\subsection{Ablation Studies}

\subsubsection{Camera Count}
\begin{figure}[pos=!tbp]
\centering
\includegraphics[width=\linewidth]{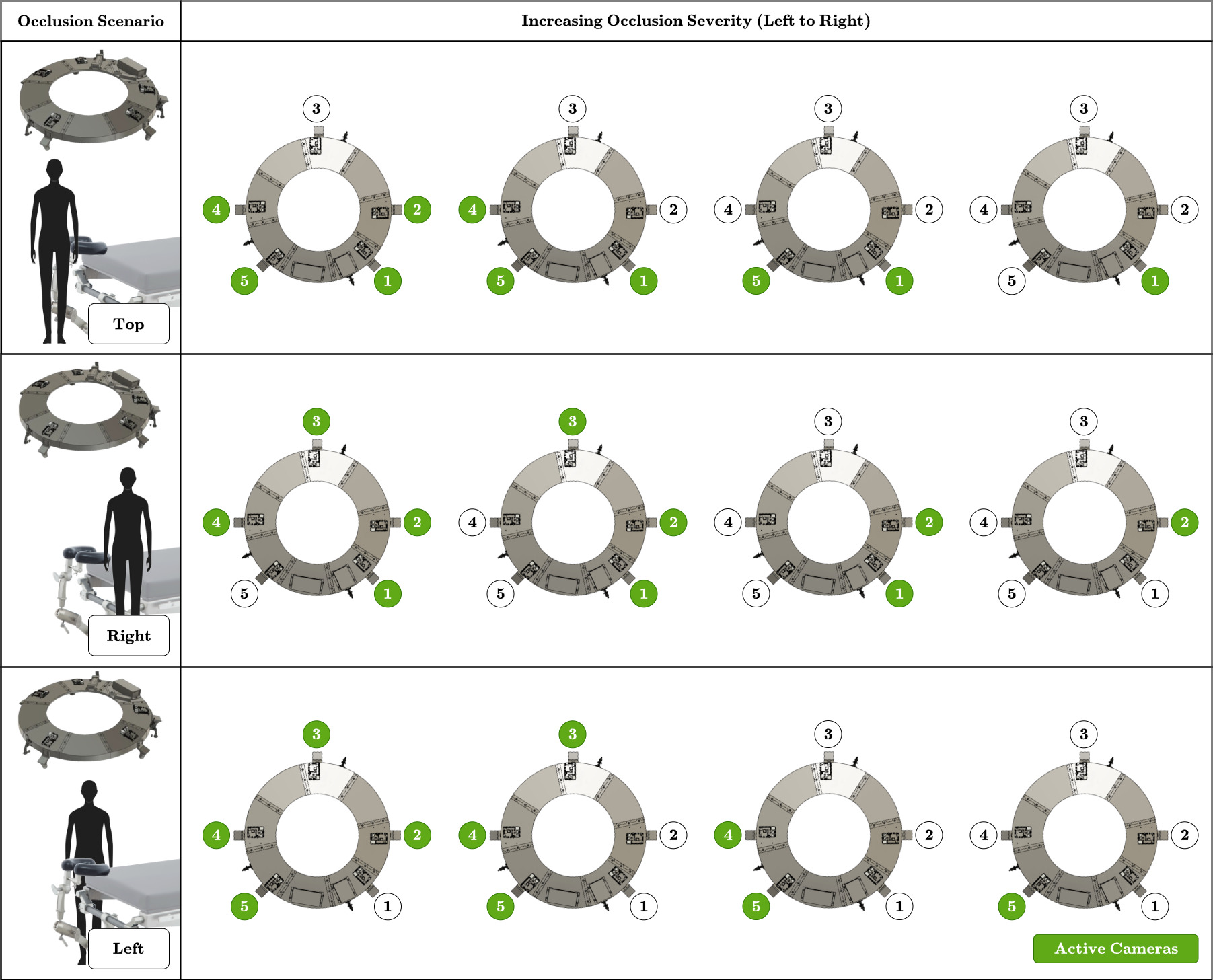}
\caption{Head phantom experiment occlusion scenarios and severity scale.}
\label{fig:configs}
\end{figure}

To evaluate robustness to camera-level occlusion, we perform an ablation study over camera subsets. For the grid experiment, we evaluate all possible camera combinations for comprehensive performance analysis. This includes all $\sum_{r=1}^{N_{\text{cams}}} \binom{N_{\text{cams}}}{r} = 2^{N_{\text{cams}}} - 1 = 31$ combinations for $N_{\text{cams}} = 5$, enabling an exhaustive assessment of how camera count and configuration affect tracking accuracy. For the head phantom experiment, we simulate realistic clinical occlusion patterns by defining three specific scenarios (top, left, and right) relative to the operating table, where clinicians or instruments would typically obstruct the field of view. We categorize occlusion severity into five levels, ranging from severe, with only one active camera, to no occlusion, with all cameras active. Figure~\ref{fig:configs} illustrates these occlusion patterns. Results are aggregated across all trials for the grid experiment across all 31 combinations, and for the head phantom across each occlusion scenario and severity level. Detailed per-scenario data for the head phantom are provided in Table~\ref{tab:head_results_combined_occlusion_scenarios}. Table~\ref{tab:results} presents overall performance for the grid and head phantom experiments respectively.

\subsubsection{Filtering Method}
To assess the benefits of adaptive noise modeling, an ablation study is performed comparing the system's performance under three configurations: 1) the main implementation using adaptive Kalman filtering for both multi-view monocular pose fusion and multi-view triangulation tracking modes, 2) an implementation with a non-adaptive Kalman filter (see Appendix~\ref{sec:noise_profiling}), and 3) an implementation without any filter (where poses in monocular pose fusion are averaged across cameras).

\section{Results}\label{sec:results}

Results for all experiments are summarized in Table~\ref{tab:results}. For the grid experiment, each RMSE value represents the aggregation across all the possible camera combinations for that number of cameras. For the head phantom experiment, each RMSE value represents the aggregation across the three occlusion scenarios. One trial with a specific camera configuration results in a single RMSE value. The tables report the median and standard deviation over all RMSE experiments.

\begin{table*}[width=\FullWidth,pos=!tbp]
\centering
\caption{Registration RMSE in mm (median ± standard deviation) across 20 trials for Grid and Head Phantom experiments. Total trajectory RMSE pools valid samples across the measured curves for both systems.}
\resizebox{\textwidth}{!}{
\begin{tabular}{lc*{7}{c}}
\toprule
& \multirow{2}{*}{\makecell[c]{\# of Cameras}} 
  & \multicolumn{3}{c}{Monocular Pose Fusion} 
  & \multicolumn{3}{c}{Triangulation} 
  & \multirow{2}{*}{fusionTrack 500} \\
\cmidrule(lr){3-5} \cmidrule(lr){6-8}
& & Basic & Kalman & Kalman Adaptive & Basic & Kalman & Kalman Adaptive & \\
\midrule
\multicolumn{9}{l}{\textbf{Grid Experiment}} \\
\midrule
\multirow{5}{*}{\makecell{Fiducial Point\\Median Registration RMSE}}
& 1 & 1.2224 (±0.2577) & 1.1693 (±0.2532) & 1.1732 (±0.2531) & N/A & N/A & N/A & \multirow{5}{*}{0.5410 (±0.0420)} \\
& 2 & 0.9697 (±0.1757) & 0.9704 (±0.1734) & 1.0193 (±0.1728) & 0.9942 (±0.3259) & 0.9871 (±0.3241) & 0.9648 (±0.3158) & \\
& 3 & 0.8419 (±0.1454) & 0.8576 (±0.1445) & 0.9174 (±0.1448) & 1.0433 (±0.4894) & 1.0427 (±0.4755) & 1.0250 (±0.4076) & \\
& 4 & 0.7683 (±0.1275) & 0.7820 (±0.1272) & 0.8517 (±0.1307) & 0.9461 (±0.2039) & 0.9487 (±0.2021) & 0.9249 (±0.1890) & \\
& 5 & 0.7189 (±0.1148) & 0.7228 (±0.1145) & 0.7995 (±0.1270) & 0.8307 (±0.1153) & 0.8271 (±0.1150) & 0.8341 (±0.1102) & \\
\cmidrule{1-9}
\multirow{5}{*}{\makecell{Target Point\\Median Registration RMSE}}
& 1 & 1.4864 (±0.3216) & 1.4287 (±0.3273) & 1.4243 (±0.3317) & N/A & N/A & N/A & \multirow{5}{*}{0.6298 (±0.0551)} \\
& 2 & 1.1121 (±0.1790) & 1.1110 (±0.1753) & 1.1504 (±0.1640) & 1.0171 (±0.3414) & 1.0146 (±0.3379) & 0.9584 (±0.3147) & \\
& 3 & 0.9568 (±0.1323) & 0.9697 (±0.1314) & 1.0104 (±0.1195) & 1.0429 (±0.6050) & 1.0396 (±0.5870) & 1.0000 (±0.5187) & \\
& 4 & 0.8690 (±0.1102) & 0.8843 (±0.1121) & 0.8936 (±0.0977) & 0.9459 (±0.1786) & 0.9367 (±0.1751) & 0.9030 (±0.1597) & \\
& 5 & 0.7855 (±0.0979) & 0.8287 (±0.1010) & 0.8290 (±0.0863) & 0.7622 (±0.0540) & 0.7588 (±0.0525) & \textbf{0.7477 (±0.0513)} & \\
\cmidrule{1-9}
\multirow{5}{*}{\makecell{Target Trajectory\\Median Registration RMSE}}
& 1 & 1.3576 (±0.3371) & 1.1840 (±0.3143) & 1.1630 (±0.3080) & N/A & N/A & N/A & \multirow{5}{*}{0.6688 (±0.0539)} \\
& 2 & 1.0589 (±0.2288) & 0.9819 (±0.2070) & 0.9595 (±0.1665) & 1.1602 (±0.3665) & 1.1403 (±0.3597) & 1.0013 (±0.3292) & \\
& 3 & 0.9237 (±0.1903) & 0.8788 (±0.1708) & 0.8590 (±0.1334) & 1.6562 (±1.0454) & 1.6334 (±1.0325) & 1.4215 (±0.9682) & \\
& 4 & 0.8659 (±0.1707) & 0.8109 (±0.1521) & 0.7895 (±0.1161) & 1.3336 (±0.5978) & 1.3176 (±0.5955) & 1.1953 (±0.5717) & \\
& 5 & 0.8100 (±0.1569) & 0.7705 (±0.1400) & \textbf{0.7675 (±0.1035)} & 1.1109 (±0.2975) & 1.1038 (±0.2953) & 1.0314 (±0.2777) & \\
\midrule
\multicolumn{9}{l}{\textbf{Head Phantom Experiment}} \\
\midrule
\multirow{5}{*}{\makecell{Fiducial Point\\Median Registration RMSE}}
& 1 & 1.5134 (±0.3488) & 1.4446 (±0.3284) & 1.4349 (±0.3264) & N/A & N/A & N/A & \multirow{5}{*}{0.8462 (±0.0532)} \\
& 2 & 1.2624 (±0.2279) & 1.2429 (±0.2198) & 1.2272 (±0.1796) & 1.2363 (±0.2821) & 1.2355 (±0.2857) & 1.1997 (±0.2953) & \\
& 3 & 1.0766 (±0.1450) & 1.0802 (±0.1532) & 0.9985 (±0.1337) & 1.2541 (±0.2772) & 1.2551 (±0.2742) & 1.2458 (±0.2671) & \\
& 4 & 1.0276 (±0.1218) & 1.0375 (±0.1320) & 0.9348 (±0.1080) & 1.1028 (±0.1022) & 1.0978 (±0.0997) & 1.0787 (±0.0933) & \\
& 5 & 0.9952 (±0.1042) & 1.0332 (±0.1047) & 0.8973 (±0.0977) & 1.0695 (±0.0735) & 1.0686 (±0.0715) & 1.0687 (±0.0649) & \\
\cmidrule{1-9}
\multirow{5}{*}{\makecell{Target Point\\Median Registration RMSE}}
& 1 & 2.8598 (±1.2893) & 2.7622 (±1.2042) & 2.7304 (±1.2023) & N/A & N/A & N/A & \multirow{5}{*}{1.0879 (±0.1140)} \\
& 2 & 2.0171 (±0.8254) & 2.0813 (±0.8253) & 2.0716 (±0.5813) & 1.7860 (±0.5421) & 1.7761 (±0.5373) & 1.7163 (±0.5355) & \\
& 3 & 1.8345 (±0.6181) & 1.9141 (±0.6325) & 1.6791 (±0.4441) & 1.6027 (±0.5950) & 1.6059 (±0.5882) & 1.6082 (±0.5634) & \\
& 4 & 1.7375 (±0.5295) & 1.8053 (±0.5511) & 1.5535 (±0.3524) & 1.4257 (±0.2897) & 1.4312 (±0.2868) & 1.3957 (±0.2834) & \\
& 5 & 1.5759 (±0.4657) & 1.6535 (±0.5066) & 1.4813 (±0.3523) & 1.2997 (±0.1885) & 1.2934 (±0.1897) & \textbf{1.2774 (±0.1938)} & \\
\cmidrule{1-9}
\multirow{5}{*}{\makecell{Target Trajectory\\Median Registration RMSE}}
& 1 & 2.4384 (±0.8503) & 2.4414 (±0.8697) & 2.4446 (±0.8690) & N/A & N/A & N/A & \multirow{5}{*}{0.7475 (±0.0948)} \\
& 2 & 1.5667 (±0.6589) & 1.6503 (±0.6855) & 1.5826 (±0.6055) & 1.4446 (±0.4989) & 1.4290 (±0.5039) & 1.3458 (±0.5034) & \\
& 3 & 2.0989 (±2.3832) & 2.2846 (±2.0119) & 1.5627 (±0.4856) & 1.5059 (±0.6478) & 1.5000 (±0.6472) & 1.4609 (±0.6320) & \\
& 4 & 1.9652 (±1.5974) & 2.1550 (±1.4467) & 1.5075 (±0.3575) & 1.2563 (±0.4529) & 1.2563 (±0.4484) & 1.2193 (±0.4420) & \\
& 5 & 2.8220 (±1.2112) & 2.6056 (±1.1142) & 1.4894 (±0.3249) & 1.1245 (±0.1549) & \textbf{1.1151 (±0.1551)} & 1.1197 (±0.1641) & \\
\bottomrule
\end{tabular}
}
\label{tab:results}
\end{table*}

\begin{figure*}[width=\textwidth,pos=!tbp]
  \centering

  \begin{subfigure}{0.48\linewidth}
    \centering
    \includegraphics[width=\linewidth]{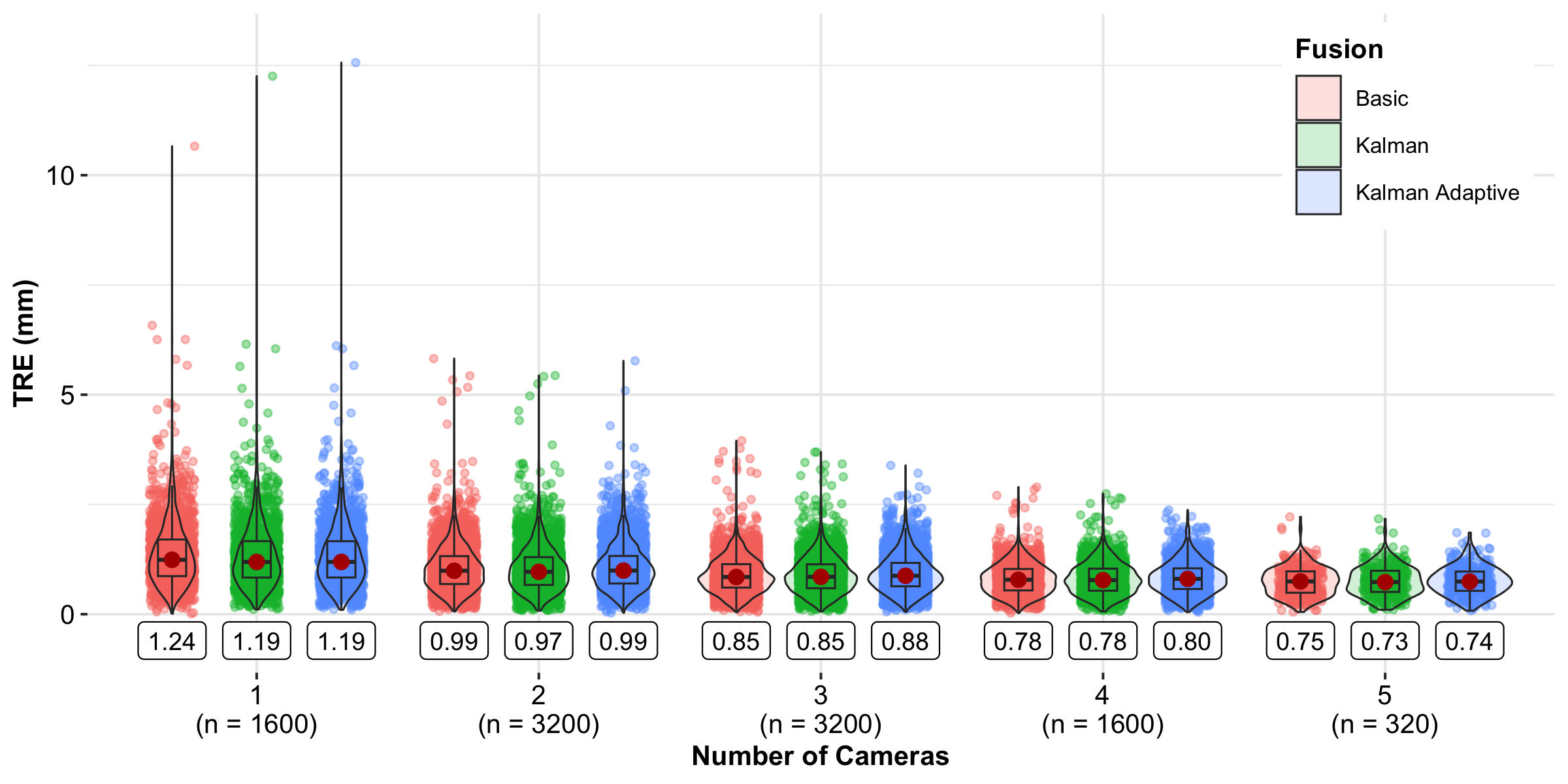}
    \caption{Point TRE – Monocular Pose Fusion}
    \label{fig:pt_tre_mono}
  \end{subfigure}
  \hfill
  \begin{subfigure}{0.48\linewidth}
    \centering
    \includegraphics[width=\linewidth]{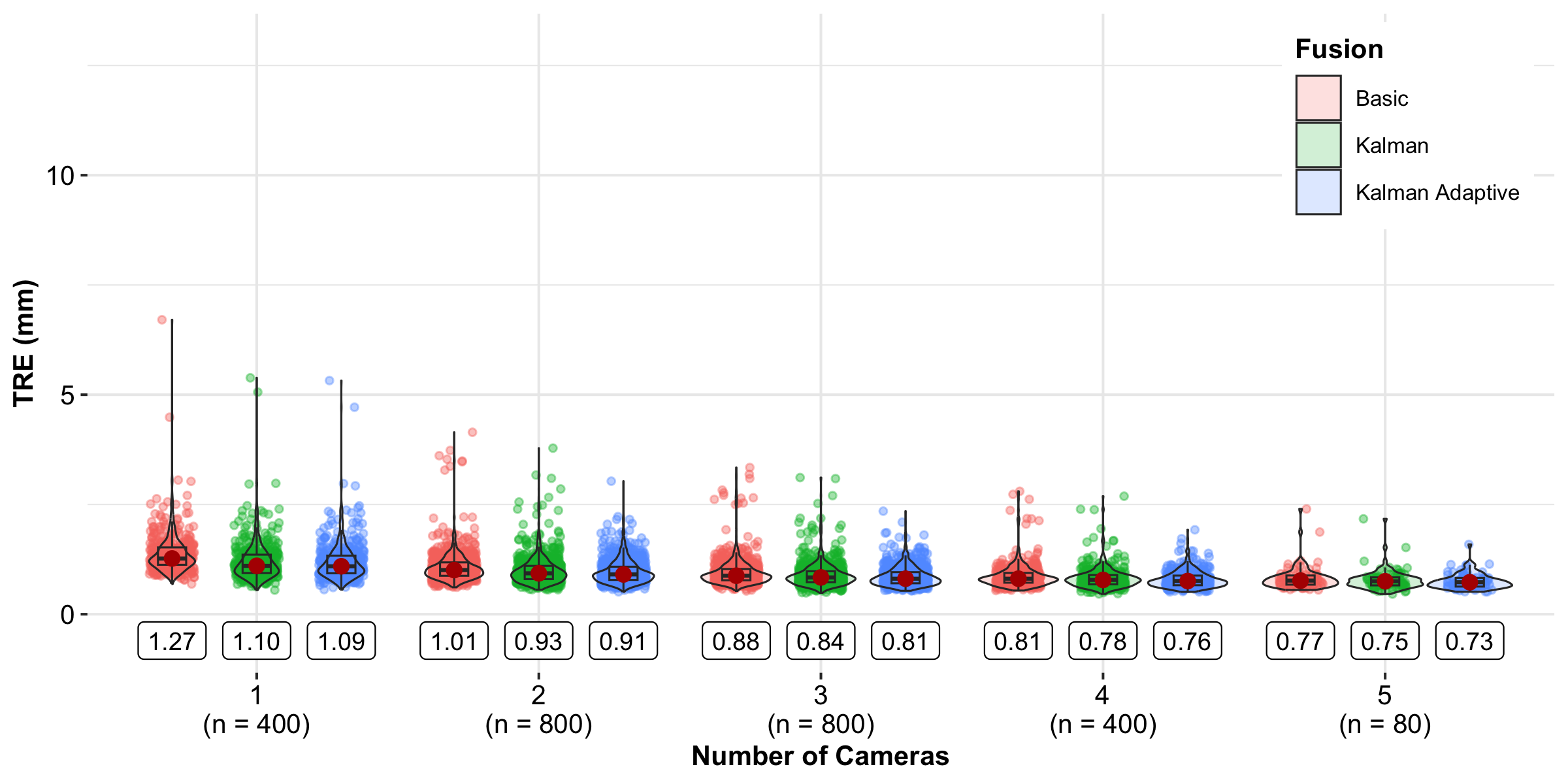}
    \caption{Trajectory TRE – Monocular Pose Fusion}
    \label{fig:traj_tre_mono}
  \end{subfigure}

  \vspace{0.8em}

  \begin{subfigure}{0.48\linewidth}
    \centering
    \includegraphics[width=\linewidth]{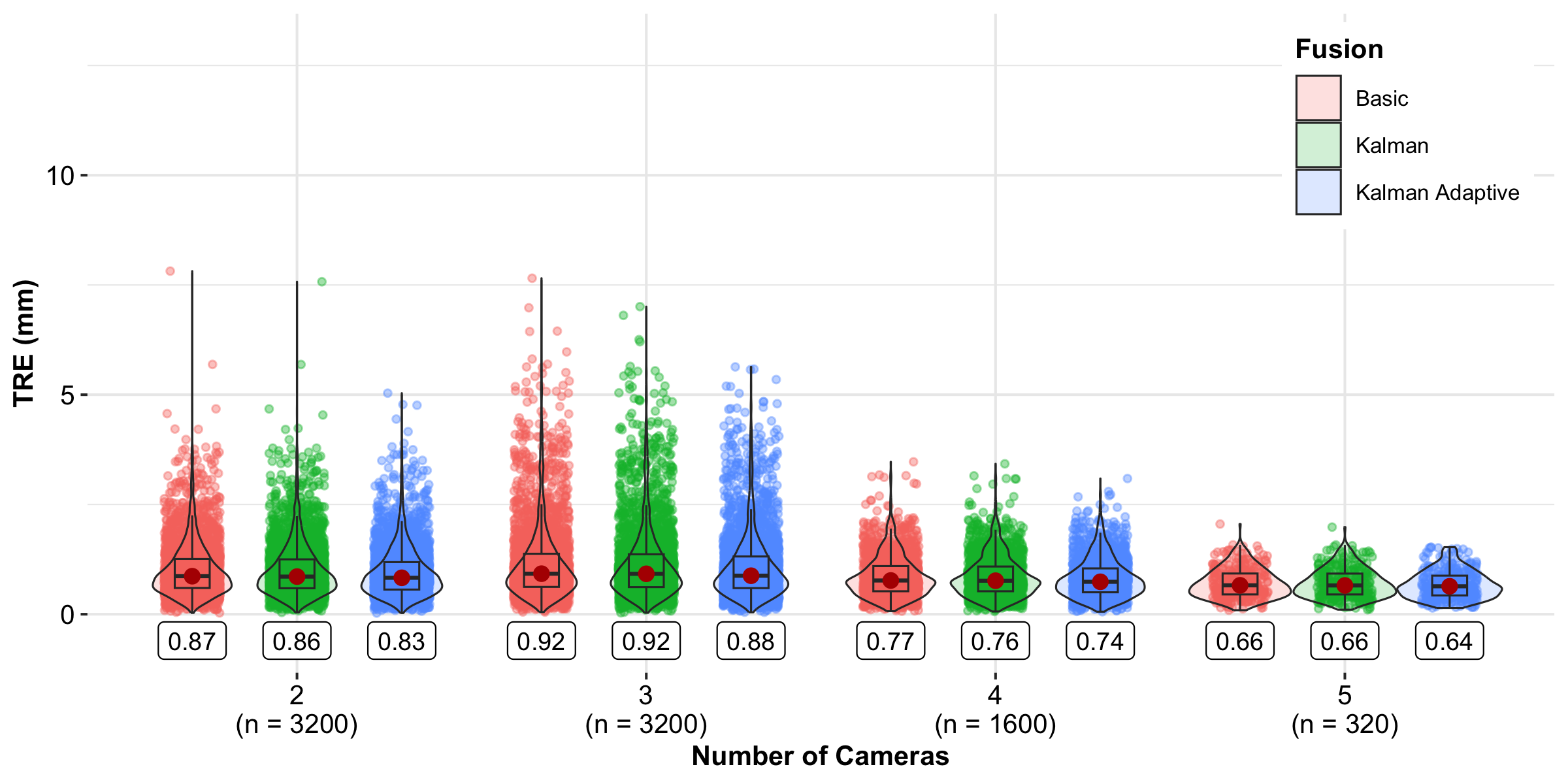}
    \caption{Point TRE – Triangulation}
    \label{fig:pt_tre_triang}
  \end{subfigure}
  \hfill
  \begin{subfigure}{0.48\linewidth}
    \centering
    \includegraphics[width=\linewidth]{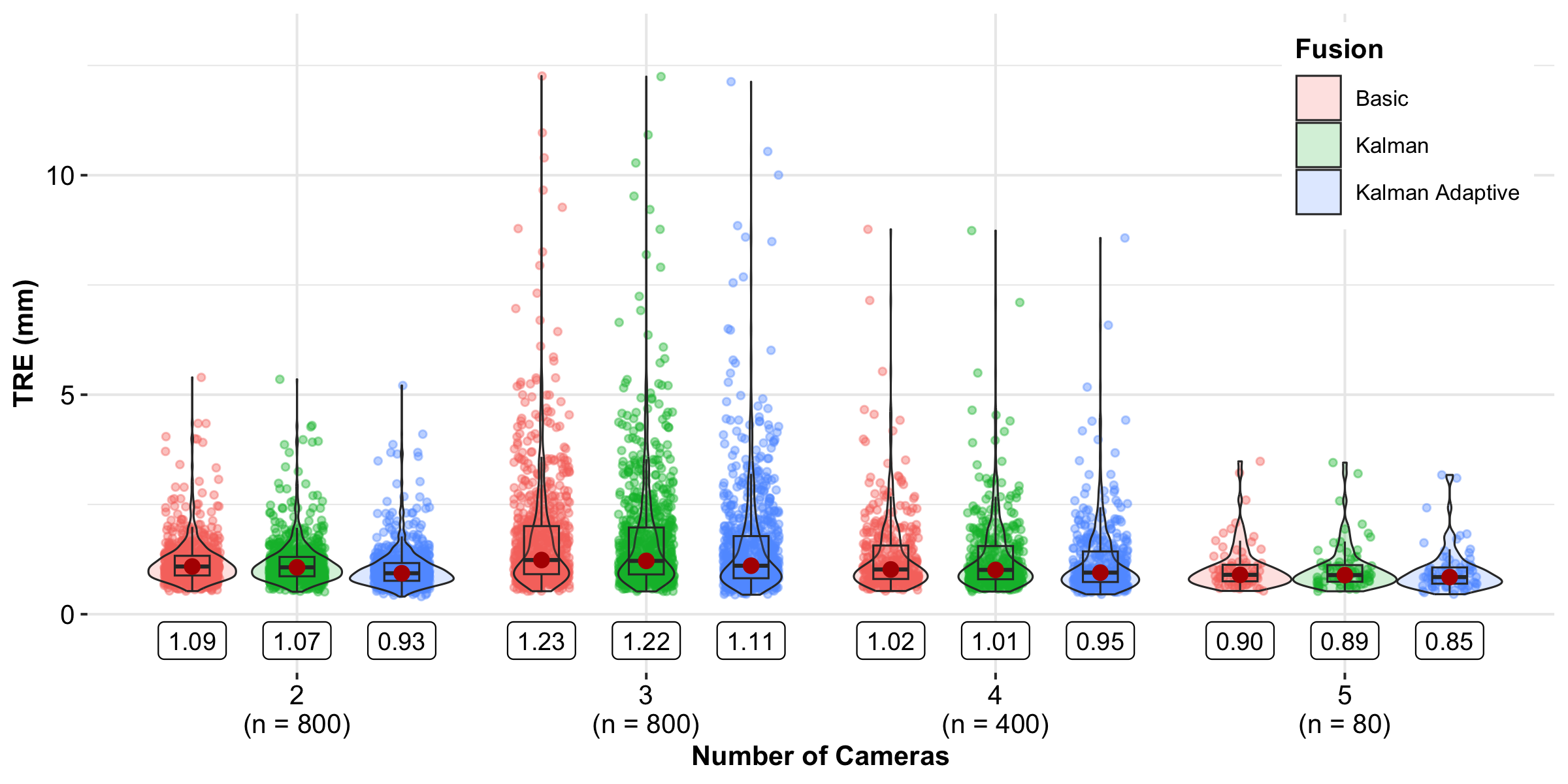}
    \caption{Trajectory TRE – Triangulation}
    \label{fig:traj_tre_triang}
  \end{subfigure}

  \caption{TRE analysis on the machined grid showing filtering method comparison across camera counts. 
           All rows show individual point TRE values (left column) and per-trajectory RMSE values (right column).
           \textbf{Top row:} Monocular pose fusion. 
           \textbf{Bottom row:} Triangulation.
           Boxes = IQR, red dots = median, violins = kernel density. Statistical comparisons use 20 paired trial summaries.}
  \label{fig:combined_tre}
\end{figure*}

\begin{figure}[pos=!tbp]
  \centering
  \includegraphics[width=\linewidth]{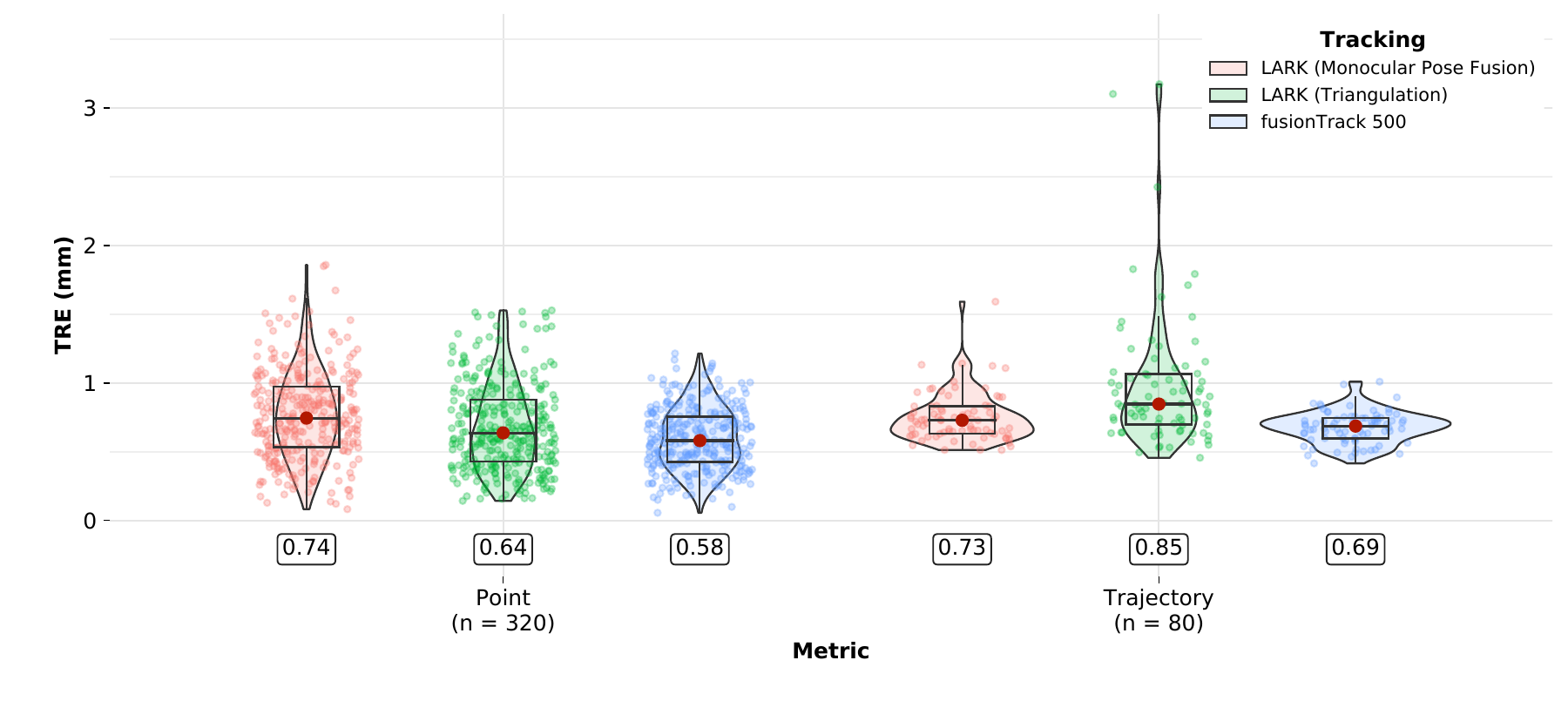}
  \caption{Five-camera performance comparison with adaptive Kalman filtering on the machined grid. Both LARK pose estimation methods (monocular pose fusion and triangulation) are compared against the gold-standard fusionTrack 500 for point localization (left) and trajectory tracking (right). Boxes = IQR, red dots = median, violins = kernel density.}
  \label{fig:grid_comparison}
\end{figure}

\begin{figure*}[width=\textwidth,pos=!tbp]
  \centering

  \begin{subfigure}{0.48\linewidth}
    \centering
    \includegraphics[width=\linewidth]{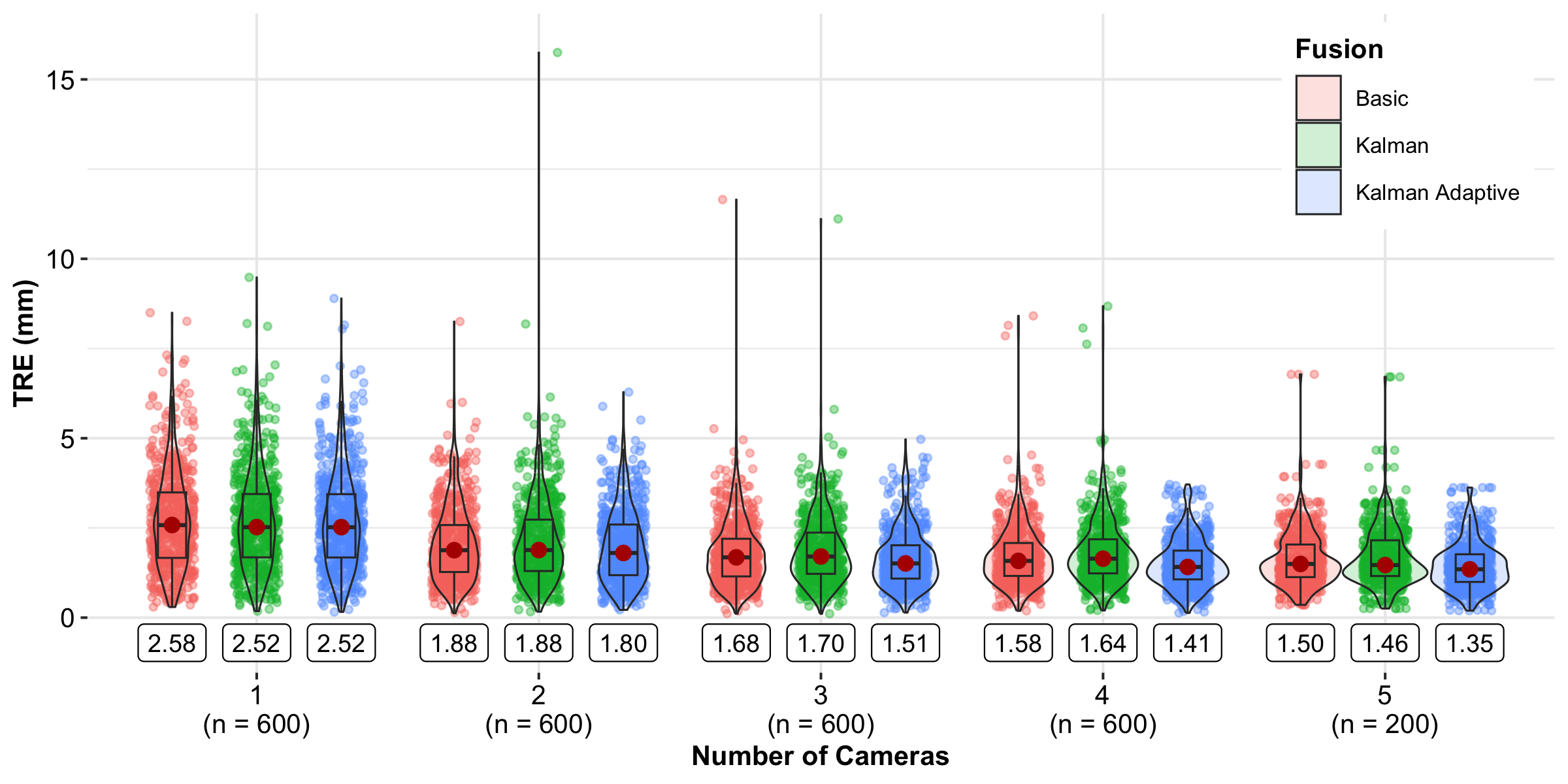}
    \caption{Point TRE – Monocular Pose Fusion}
    \label{fig:pt_tre_mono_head}
  \end{subfigure}
  \hfill
  \begin{subfigure}{0.48\linewidth}
    \centering
    \includegraphics[width=\linewidth]{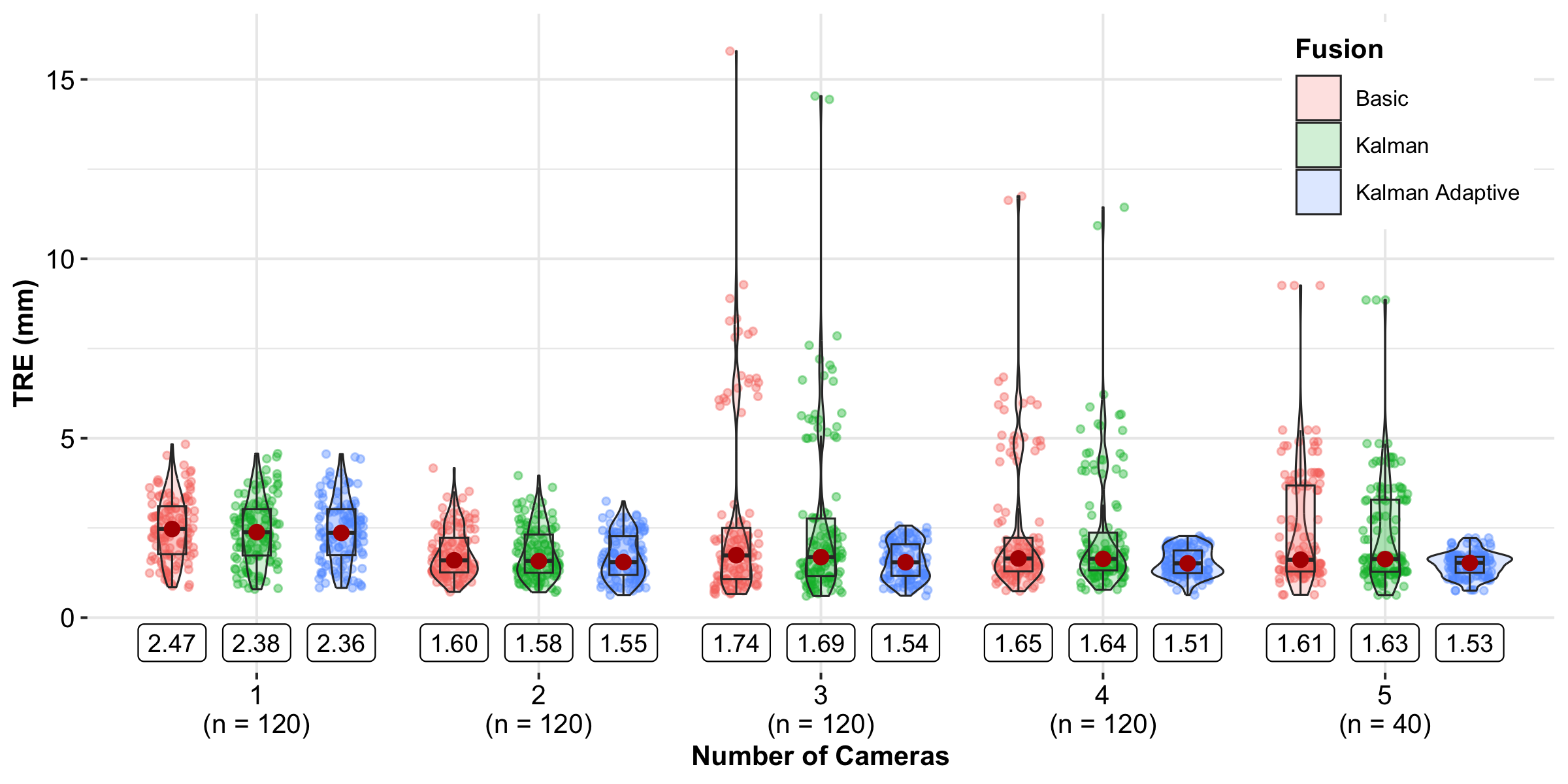}
    \caption{Trajectory TRE – Monocular Pose Fusion}
    \label{fig:traj_tre_mono_head}
  \end{subfigure}

  \vspace{0.8em}

  \begin{subfigure}{0.48\linewidth}
    \centering
    \includegraphics[width=\linewidth]{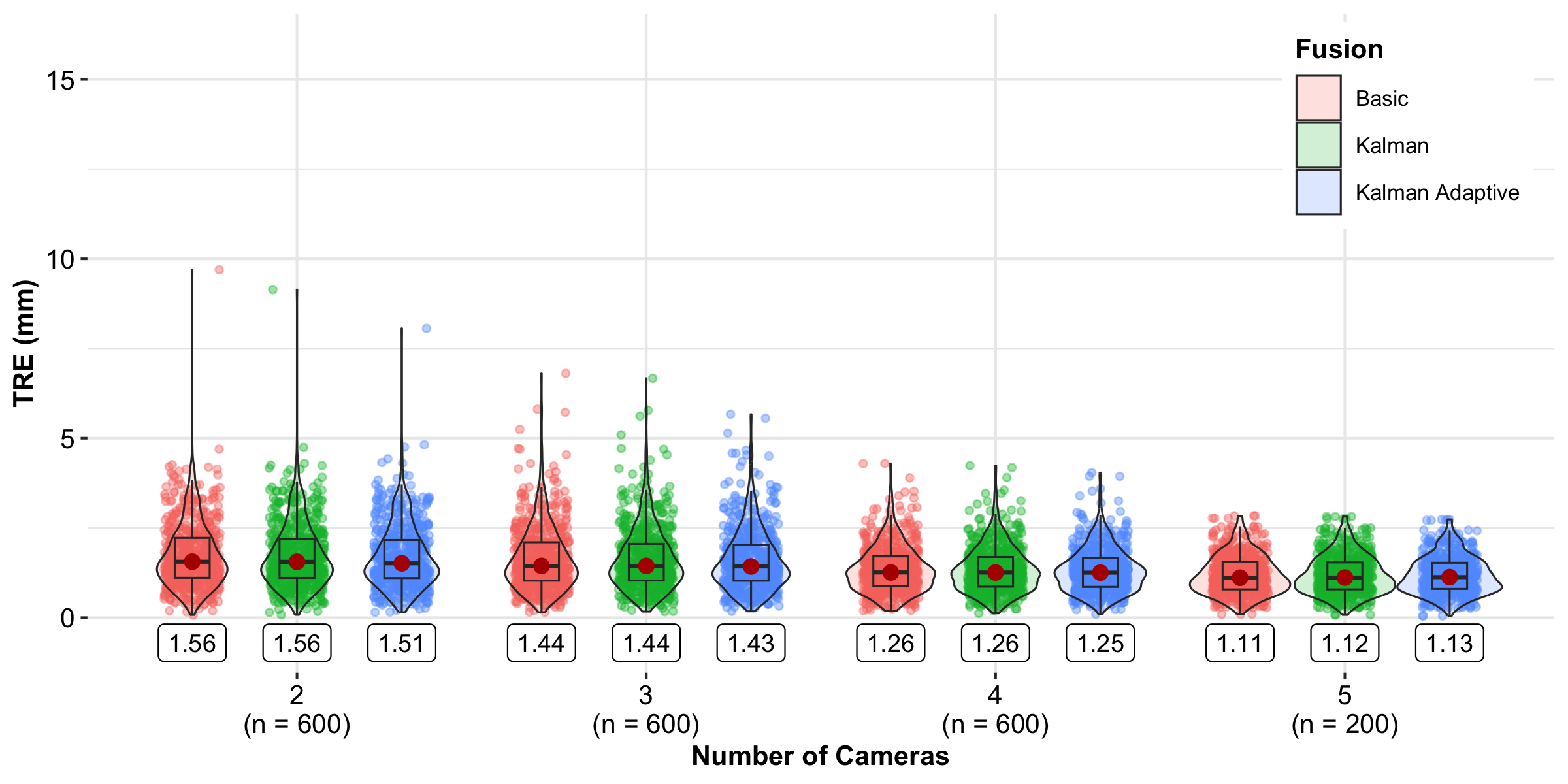}
    \caption{Point TRE – Triangulation}
    \label{fig:pt_tre_triang_head}
  \end{subfigure}
  \hfill
  \begin{subfigure}{0.48\linewidth}
    \centering
    \includegraphics[width=\linewidth]{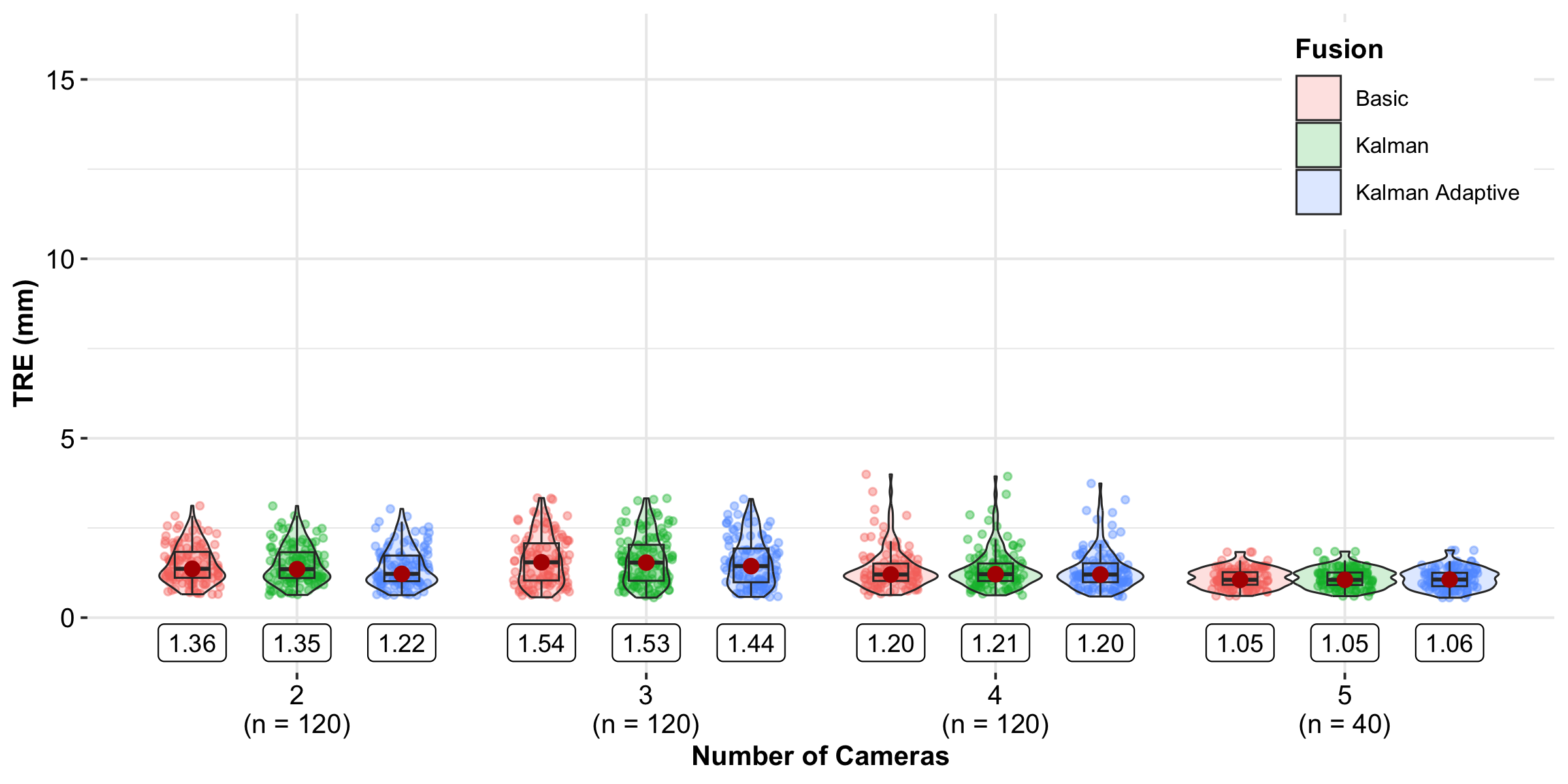}
    \caption{Trajectory TRE – Triangulation}
    \label{fig:traj_tre_triang_head}
  \end{subfigure}

  \caption{TRE analysis on the SLA-printed head phantom showing filtering method comparison across camera counts. 
           All rows show individual point TRE values (left column) and per-trajectory RMSE values (right column).
           \textbf{Top row:} Monocular pose fusion. 
           \textbf{Bottom row:} Triangulation.
           Boxes = IQR, red dots = median, violins = kernel density. Statistical comparisons use 20 paired trial summaries.}
  \label{fig:combined_tre_head}
\end{figure*}

Table~\ref{tab:results} summarizes trial- and configuration-level RMSE values. In Figures~\ref{fig:combined_tre}--\ref{fig:combined_tre_head}, the point TRE distributions contain individual target-point errors, whereas the trajectory TRE distributions contain one RMSE value per measured trajectory for each trial and camera configuration. The plotted medians therefore summarize different units of observation for point and trajectory tracking. For the respective FRE plots, see Figure~\ref{fig:fre_quad}.

\begin{figure}[pos=!tbp]
  \centering
  \includegraphics[width=\linewidth]{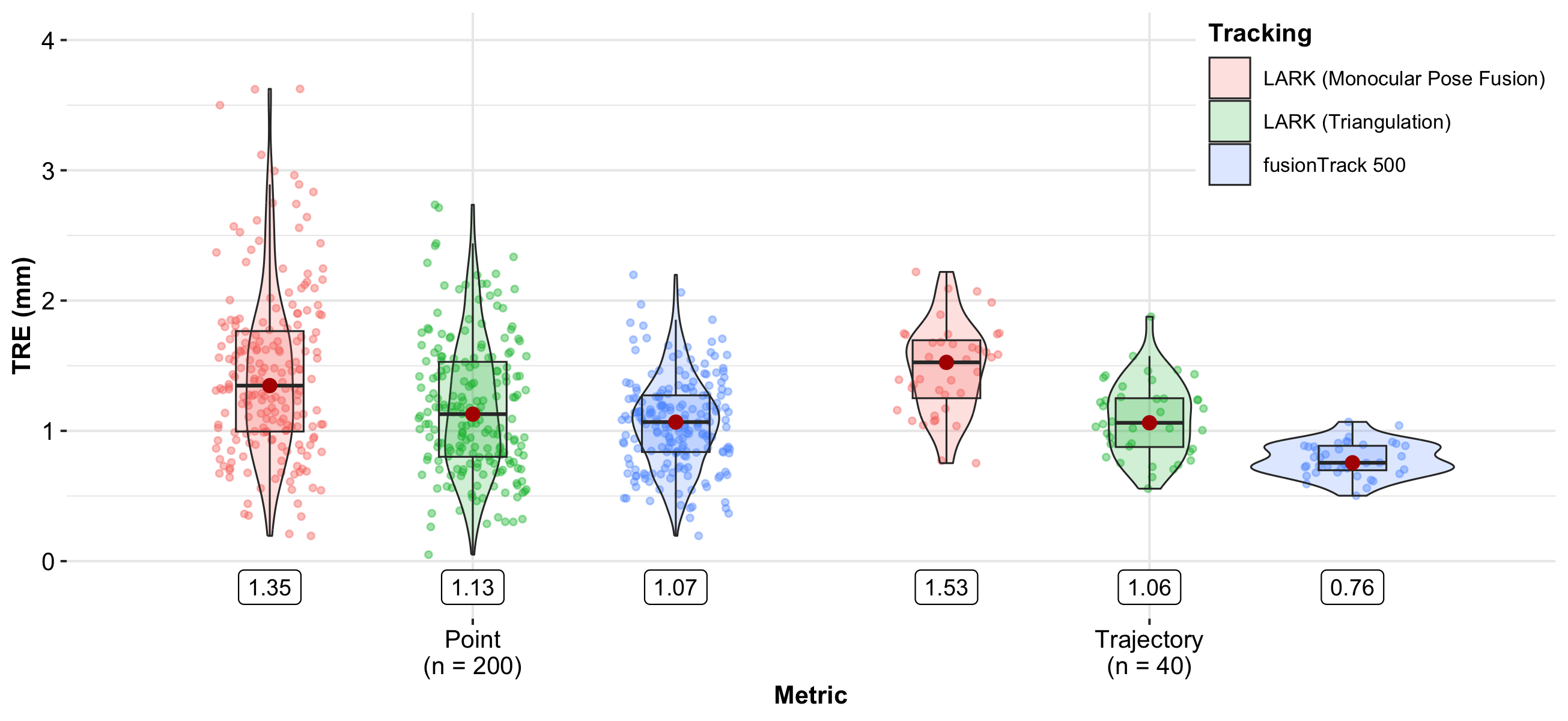}
  \caption{Five-camera performance comparison with adaptive Kalman filtering on the anatomical head phantom. Under challenging viewing conditions with partial marker occlusion, triangulation outperforms monocular pose fusion for both point localization (1.13~mm versus 1.35~mm) and trajectory tracking (1.06~mm versus 1.53~mm). Boxes = IQR, red dots = median, violins = kernel density.}
  \label{fig:head_comparison}
\end{figure}

Statistical comparisons use the 20 acquisition trials per phantom as independent units. Within each trial and tracking setting, squared point errors or squared per-trajectory RMSEs are averaged within each camera configuration; configuration mean squares are then averaged equally and square-rooted. The five-camera head measurements are included once across occlusion scenarios. For pooled camera-count comparisons, mean squares are additionally averaged equally across available tracking methods and filters within each trial (pose fusion alone with one camera; both methods with two to five cameras). We use two-sided paired Wilcoxon signed-rank tests with exact sign-assignment probabilities and Holm correction ($\alpha=0.05$): three filter comparisons per experiment, camera count, tracking method, and metric, or ten camera-count comparisons per experiment and metric. Paired sign tests assess sensitivity to the signed-rank symmetry assumption; Tables~\ref{tab:pairwise_pvals_grid_head}--\ref{tab:pairwise_methods_head} mark comparisons whose significance decisions differ between the tests.

Pooled trial-summary errors generally decrease with more cameras, although improvements are not monotonic. Compared with the basic method, adaptive Kalman filtering significantly reduces pose-fusion trajectory error at all camera counts on the grid and at three to five cameras on the head phantom. It also significantly reduces triangulation trajectory error at all grid camera counts. Both paired tests support these filtering comparisons.

\begin{figure}[pos=!tbp]
  \centering
  \includegraphics[width=\linewidth]{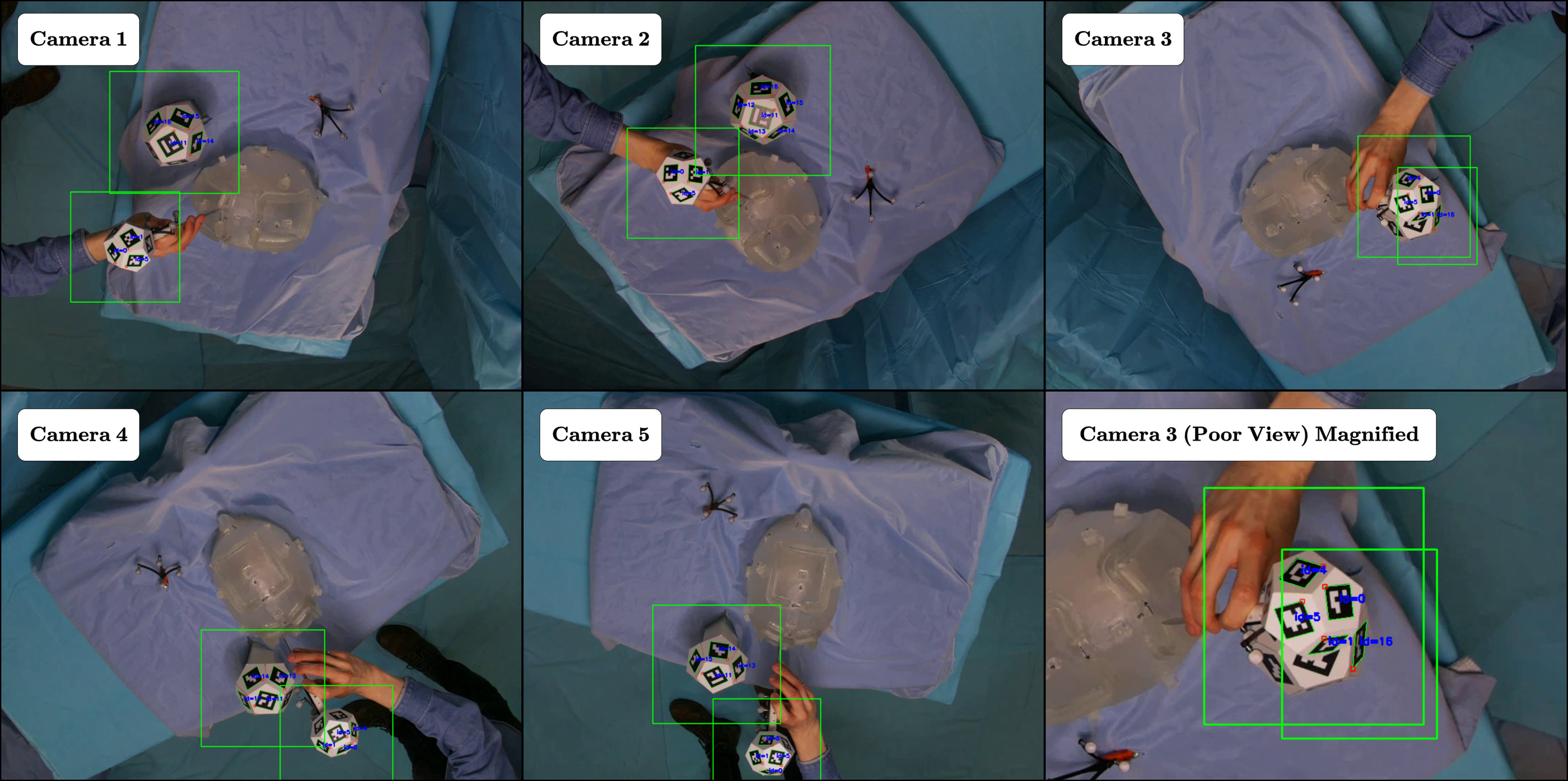}
  \caption{Camera views during trajectory tracking at the moment of maximum error in the head phantom experiment. All five camera views are shown with detected markers highlighted in green boxes. Camera 3 (top right) demonstrates a poor view condition where the pointer partially occludes the reference, leaving only a single marker visible at an oblique angle to the image plane. The bottom right panel shows a magnified view of camera 3's compromised perspective.}
  \label{fig:camera_views_poor}
\end{figure}

\begin{figure*}[width=\textwidth,pos=!tbp]
  \centering
  \includegraphics[width=\linewidth]{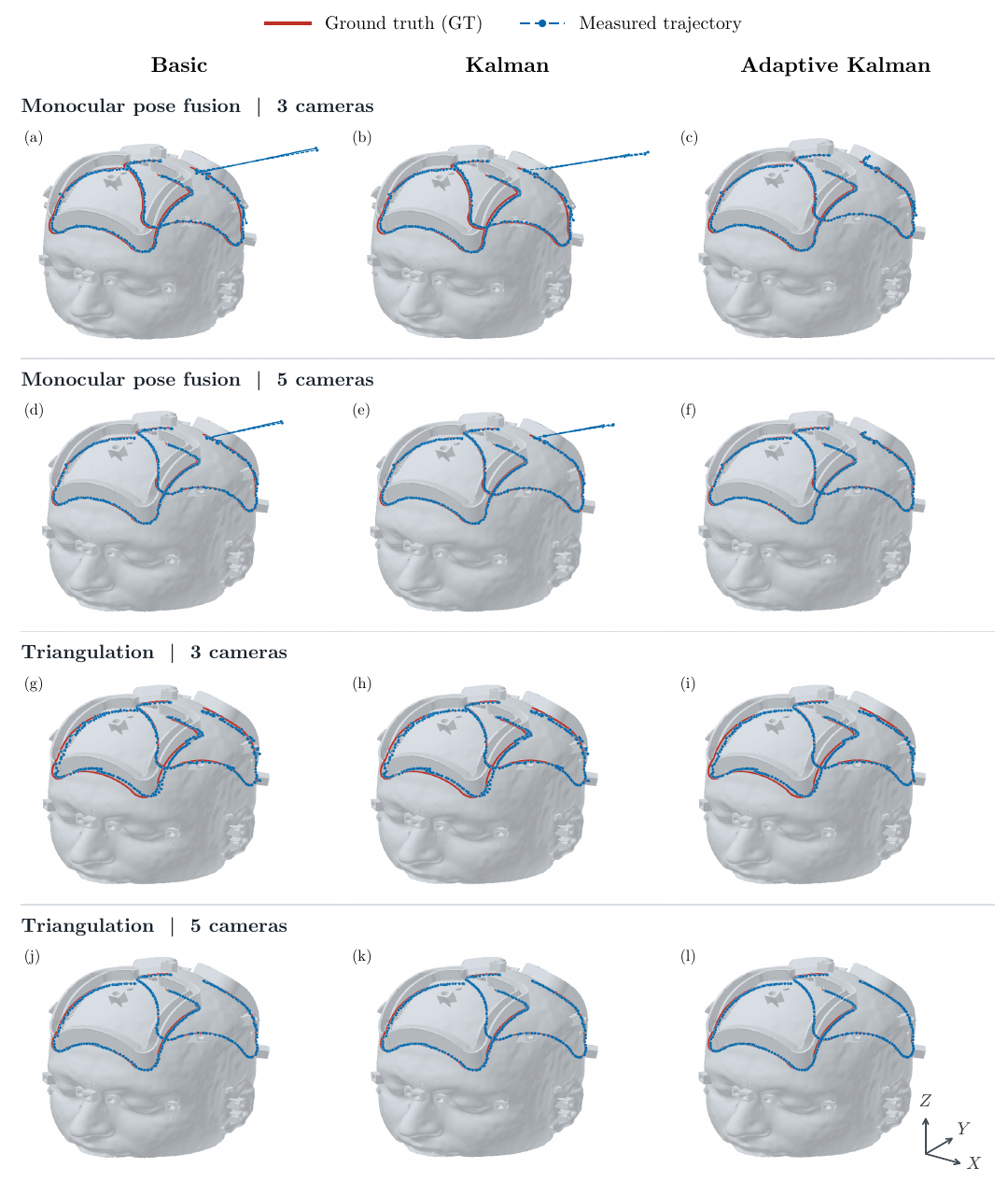}
  \caption{Comparison of tracked trajectories versus ground truth for the left occlusion scenario in the head phantom experiment trial which produced the largest overall outlier. Top two rows show monocular pose fusion for 3-camera and 5-camera configurations respectively. Bottom two rows show triangulation for 3-camera and 5-camera configurations respectively. Columns from left to right show Basic, Kalman, and Adaptive Kalman filtering. Red lines represent ground truth, blue connected dots represent tracked trajectories. The cropped, shaded head provides qualitative spatial context; the shared XYZ triad indicates orientation only. Panel scales differ, so apparent trajectory sizes should not be compared across panels.}
  \label{fig:trajectory_comparison_left_occlusion}
\end{figure*}

To illustrate the mechanisms underlying these performance differences, Figures~\ref{fig:camera_views_poor} and \ref{fig:trajectory_comparison_left_occlusion} examine the worst-case trial from the head phantom experiment, where camera 3 observes only a single marker at an oblique angle and tracking errors accumulate towards the trajectory's end. Section~\ref{sec:discussion} analyses these contrasting behaviors in detail.

\section{Discussion}\label{sec:discussion}

The grid and head phantom experiments reveal distinct performance differences between multi-view monocular pose fusion and triangulation that depend critically on viewing conditions and scene geometry.

On the machined grid, where all cameras maintain reasonable multi-marker views, both tracking methods achieve sub-millimeter accuracy with five cameras using adaptive Kalman filtering. Multi-view monocular pose fusion achieves 0.74~mm median point TRE while triangulation attains 0.64~mm. Figure~\ref{fig:grid_comparison} directly compares both methods against the gold-standard fusionTrack 500 at the five-camera configuration with adaptive Kalman filtering, showing comparable point localization accuracy.

The methods differ markedly in their response to camera count. Monocular pose fusion generally improves as cameras are added, with particularly strong performance in trajectory tracking (0.73~mm TRE with five cameras and adaptive Kalman filtering; Figure~\ref{fig:combined_tre}b). In this method, per-camera covariance weighting adjusts each pose estimate's contribution to the fused result. Triangulation depends on inter-camera extrinsic calibration, whereas monocular pose fusion estimates the pointer-to-reference pose separately in each camera. Both methods combine observations across cameras and can be affected by inter-camera timing errors during motion. On the grid, monocular pose fusion achieves lower median trajectory TRE than triangulation with five cameras and adaptive filtering.

Triangulation demonstrates unexpected non-monotonic behavior, with accuracy degrading at intermediate camera counts (1.11~mm trajectory TRE at three cameras) before recovering with five cameras (Figure~\ref{fig:combined_tre}d). This sensitivity likely arises from geometric conditioning issues when cameras are added at certain angles relative to the planar grid, affecting ray intersection quality. However, triangulation maintains superior static point localization (0.64~mm versus 0.74~mm for monocular fusion; Figure~\ref{fig:combined_tre}a,c), likely due to its explicit use of multi-view geometric constraints. The direct comparison in Figure~\ref{fig:grid_comparison} shows median trajectory TRE values of 0.73~mm for monocular pose fusion, 0.85~mm for triangulation, and 0.69~mm for the fusionTrack 500.

The clearest benefit of adaptive Kalman filtering is observed in monocular trajectory tracking, where it reduces the spread of errors (Figure~\ref{fig:combined_tre}a--d). Its effect on point localization and triangulation varies with camera count and filtering method.

The head phantom reveals fundamentally different performance characteristics. The phantom's complex three-dimensional geometry creates more challenging conditions than the planar grid: registration requires steeper viewing angles that increase marker-level occlusions, the larger spatial extent along the reference frame's z-axis challenges depth estimation, and experimental limitations include potential warping of the SLA-printed material and wear from repeated use. With five cameras and adaptive Kalman filtering, neither method achieves sub-millimeter median point or trajectory TRE on the head phantom. The fixed camera positions, while ensuring fair comparison, create a more demanding scenario than typical clinical use where cameras can be repositioned to optimize viewing angles for specific operative regions.

The critical difference emerges in how each tracking method handles severely degraded views. The occlusion scenario analysis (see Table~\ref{tab:head_results_combined_occlusion_scenarios}) reveals that camera 3 consistently produced the largest outliers when it could observe only a single marker. Figure~\ref{fig:camera_views_poor} illustrates this problematic condition where the pointer partially occludes the reference, leaving camera 3 with a compromised view worse than complete occlusion. The oblique viewing angle in this case compounds the tracking error.

In monocular pose fusion, camera 3's erroneous pose estimate remains a significant component of the fused result. The adaptive Kalman filter adjusts weights according to measurement covariance, but a single-marker pose does not always yield sufficiently high variance to trigger strong down-weighting if the detection itself appears stable. As a result, some outliers persist (Figure~\ref{fig:combined_tre_head}a,b). Even in such cases, however, the filter substantially mitigates errors compared to the basic approach. For example, in the three-camera left occlusion scenario, the adaptive Kalman filter reduces the median trajectory RMSE from 4.45$\pm$2.36~mm to 1.47$\pm$0.42~mm (see Table~\ref{tab:head_results_combined_occlusion_scenarios}).

Triangulation proves remarkably robust to the same poor view (Figure~\ref{fig:combined_tre_head}c,d). As shown in Figure~\ref{fig:trajectory_comparison_left_occlusion}, triangulation maintains stable tracking without large outliers across all filtering implementations. This behavior is consistent with the different ways the methods use sparse observations: a single marker supplies four coplanar corners, which permit PnP pose estimation but can yield ambiguous or noise-sensitive estimates. Triangulation instead combines corner observations across cameras before estimating the instrument pose. Additionally, camera 3's limited observations affect only a few corner measurements among potentially dozens from all cameras combined, diluting its impact on the final pose computation. With five cameras and adaptive Kalman filtering, triangulation outperforms monocular pose fusion on the anatomical phantom for both point localization (1.13~mm versus 1.35~mm) and trajectory tracking (1.06~mm versus 1.53~mm; Figure~\ref{fig:head_comparison}).

Adaptive Kalman filtering reduces the spread of trajectory errors in monocular pose fusion, although stable erroneous single-marker estimates can retain appreciable weight. The effect of filtering on triangulation depends on camera count and phantom.

These phantom results support further development of low-cost multi-camera tracking for image-guided navigation. The method comparison depends on viewing conditions: monocular pose fusion gives lower trajectory error on the grid, whereas triangulation gives lower point and trajectory errors on the anatomical head phantom with five cameras and adaptive filtering.

The results demonstrate the potential of commodity RGB cameras for low-cost tracking with multiple viewpoints. Three observations emerge from the comparison. First, monocular pose fusion is more affected by the illustrated poor single-marker view than triangulation. Second, adaptive Kalman filtering reduces the spread of trajectory errors in monocular pose fusion. Third, camera-subset performance depends on both the number and quality of the retained views.

Future work should explore hybrid strategies that incorporate view-quality metrics to trigger dynamic switching between tracking methods or selective camera exclusion, and combine poses derived from triangulation of stereo pairs with adaptive Kalman fusion, preventing inaccurate estimates arising from poor single-marker views while preserving the accurate trajectory tracking observed under favorable conditions.

\section{Conclusion}\label{sec:conclusion}
This work demonstrates low-cost multi-camera optical tracking using commodity RGB hardware, evaluated on a machined grid and an anatomical head phantom. Our system achieves 0.64 mm point TRE with triangulation and 0.73 mm trajectory TRE with pose fusion on a machined grid, and 1.13 mm point and 1.06 mm trajectory TRE with triangulation on an anatomical head phantom.

The two methods have complementary strengths: with five cameras and adaptive filtering, monocular pose fusion achieves lower trajectory TRE on the grid, whereas triangulation achieves lower point and trajectory TRE on the head phantom. The illustrated poor-view condition also affects triangulation less than monocular pose fusion.

The camera-subset experiments characterize how tracking accuracy changes as views become unavailable, while the poor-view example shows that the quality of retained views also matters. With tracking hardware costing under \$1,000 USD and full open-source release of designs, technical drawings, bill of materials, software, and datasets (designs and software: \url{https://nist.mni.mcgill.ca/software/}; datasets: \url{https://nist.mni.mcgill.ca/data/}), this work lowers barriers to adoption in resource-constrained settings and provides a reproducible benchmark for evaluating tracking approaches in IGS research. Clinical validation remains the critical next step to confirm translation of these laboratory results to improved surgical outcomes.%

\section*{Acknowledgements}

We acknowledge funding from the Canadian Institutes of Health Research FRN-152980, FRN-158277 and PJT-109163.

\appendix
\numberwithin{equation}{section}
\numberwithin{figure}{section}
\numberwithin{table}{section}

\section{Noise Profiling}\label{sec:noise_profiling}

To compare the performance of our adaptive Kalman filter with a constant velocity Kalman filter using static covariance matrices, we first conduct a noise profiling procedure. The procedure allows us to estimate the static measurement noise covariance matrices \( \mbf{R}_i \) for each camera and the pose block of the process noise covariance matrix \( \mbf{Q} \) for the non-adaptive filter.

The noise profiling involves collecting a series of pose measurements while the surgical instrument is held stationary at \(L\) distinct workspace locations indexed by \(\ell=1,\dots,L\). At every location we record \(T = 30\) consecutive pose estimates from each active camera
\(i\in\{1,\dots,N\}\). The sample variance of those poses gives the
location–specific measurement covariance

\begin{equation}
  \mbf{R}_{\ell,i} =
  \operatorname{Var}\!\Bigl(
    \mbf{y}_{\tau,i}\;
    \bigl|\;
    \tau = 1,\dots,T
  \Bigr),
\end{equation} where \(\mbf{y}_{\tau,i}\) is the pose returned by camera \(i\) on
frame \(\tau\) while the instrument is fixed at location \(\ell\).

We then compute the average measurement noise covariance for each camera across all sampled positions:

\begin{equation}
  \bar{\mbf{R}}_{i} \;=\;
  \frac{1}{L}\,\sum_{\ell=1}^{L}\mbf{R}_{\ell,i},\end{equation} which serves as the static measurement noise covariance matrix for camera \( i \) in the non-adaptive Kalman filter.

To formulate the static process noise covariance matrix \( \mbf{Q} \),
we average the \(\bar{\mbf{R}}_{i}\) matrices across all cameras for its
pose block and retain the same fixed velocity-noise block:

\begin{equation}
  \mbf{Q}\;=\;
  \begin{bmatrix}
    \frac{1}{N}\sum_{i=1}^{N}\bar{\mbf{R}}_{i} & \mbf{0}_{6}\\
    \mbf{0}_{6} & 0.001\,\mbf{I}_{6}
  \end{bmatrix}.
\end{equation}
Here \(N\) is the total number of cameras. Stationary noise profiling
determines the pose-noise block; the velocity variances are fixed tuning
parameters, as in the adaptive filter described in
Section~\ref{sec:mono_tracking}.

Beyond tuning the static covariance matrices for the non-adaptive Kalman filter, this procedure provides valuable insights into the spatial distribution of measurement noise within the workspace. By analyzing the norms of the covariance matrices \(\mbf{R}_{\ell,i}\), we generate heatmaps illustrating the noise profile for each camera at different positions. The Frobenius norm of each \(\mbf{R}_{\ell,i}\) matrix is computed as

\begin{equation}
  \bigl\lVert\mbf{R}_{\ell,i}\bigr\rVert_{F} \;=\;
  \sqrt{\operatorname{Tr}\!\bigl(\mbf{R}_{\ell,i}\mbf{R}_{\ell,i}^{\!\top}\bigr)}.
\end{equation}

\begin{figure}[pos=!tbp]
  \centering
  \includegraphics[width=\linewidth]{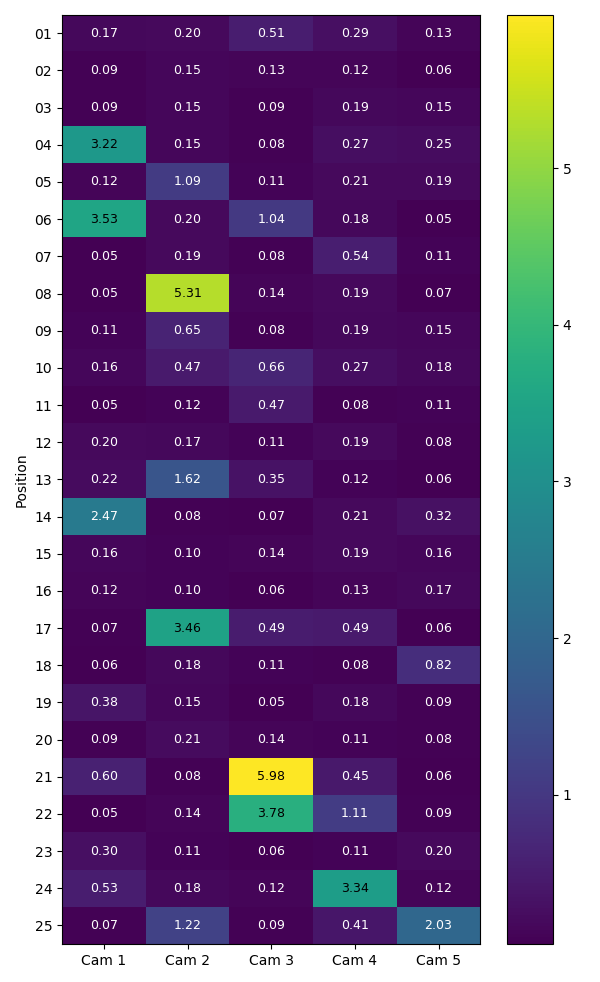}\caption{Spatial distribution of the norm of the measurement noise covariance matrices \(\mbf{R}_{\ell,i}\) for each camera across all measured positions.}
    \label{fig:R_pos}
\end{figure}

This norm provides a single scalar value to represent the overall noise level for each camera at each position. Figure \ref{fig:R_pos} shows the spatial distribution of noise across all measured positions for one of our experimental trials.

This noise profiling procedure configures the static Kalman filter using measured stationary pose variability, providing a baseline for comparison with the adaptive Kalman filter. Moreover, the insights gained can guide adjustments to lighting, camera placement, and calibration to enhance measurement precision. In the OR, this procedure could help technicians assess tracking precision during camera setup.

\section{Landmark Registration}\label{sec:landmark_registration}

Landmark registration aligns a set of measured 3-D points with their corresponding ground-truth (GT) points to determine the rigid transformation between two coordinate systems. This method is central to IGS, where precise alignment ensures that surgical instruments and patient anatomy share a consistent spatial frame of reference.

We employ a point–based registration strategy using the Orthogonal Procrustes method \citep{Fitzpatrick1998}. The aim is to find a rotation matrix \(\mbf{S}\in SO(3)\) and a translation vector \(\mbf{d}\in\mathbb{R}^3\) that minimise the difference between corresponding points in the measured and GT sets.  Suppose \(N_f\) fiducial points are available. The Fiducial Registration Error (FRE) is the root-mean-square distance between corresponding fiducials after alignment:

\begin{align}
\text{FRE} \;=\;
\sqrt{\frac{1}{N_f}\;
      \sum_{i=1}^{N_f}
      \bigl\|
        \mbf{x}_i^{\text{GT}}
        -\bigl(\mbf{S}\,\mbf{x}_i^{\text{meas}}+\mbf{d}\bigr)
      \bigr\|^{2}}
\end{align}

To determine \(\mbf{S}\) and \(\mbf{d}\), the measured and GT point sets are first translated so that their centroids lie at the origin.  Denote those centroids by \(\bar{\mbf{x}}^{\text{meas}}\) and \(\bar{\mbf{x}}^{\text{GT}}\).  After centring, the covariance matrix is

\begin{align}
\mbf{G}
\;=\;
\sum_{i=1}^{N_f}
  (\mbf{x}_i^{\text{meas}}-\bar{\mbf{x}}^{\text{meas}})
  (\mbf{x}_i^{\text{GT}}-\bar{\mbf{x}}^{\text{GT}})^{\!\top}.
\end{align}

Applying Singular Value Decomposition (SVD) to \(\mbf{G}\),

\[
\mbf{G} \;=\; \mbf{U}\,\mbs{\Sigma}\,\mbf{V}^{\!\top},
\]

the optimal rotation is

\[
\mbf{S} \;=\; \mbf{V}\,\mbf{U}^{\!\top}.
\]

If \(\det(\mbf{S})<0\) the final column of \(\mbf{V}\) is negated to prevent a reflection.  
The translation that relates the two centroids is

\[
\mbf{d} \;=\; \bar{\mbf{x}}^{\text{GT}} - \mbf{S}\,\bar{\mbf{x}}^{\text{meas}} .
\]

With \(\mbf{S}\) and \(\mbf{d}\) fixed by the fiducials, we validate the transformation on a separate set of \(N_t\) target points. The TRE is

\begin{align}
\text{TRE} \;=\;
\sqrt{\frac{1}{N_t}\;
      \sum_{j=1}^{N_t}
      \bigl\|
        \mbf{y}_j^{\text{GT}}
        -\bigl(\mbf{S}\,\mbf{y}_j^{\text{meas}}+\mbf{d}\bigr)
      \bigr\|^{2}} .
\end{align}

The same transformation can be applied to a measured trajectory  
\(\{\mbf{p}_1^{\text{meas}},\dots,\mbf{p}_{N_p}^{\text{meas}}\}\) to obtain

\[
\mbf{p}_i^{\text{trans}} \;=\; \mbf{S}\,\mbf{p}_i^{\text{meas}} + \mbf{d},
\qquad i=1,\dots,N_p .
\]

For each transformed point \(\mbf{p}_i^{\text{trans}}\) we locate its nearest neighbour \(\mbf{q}_i^{\text{GT}}\) in the GT trajectory \(\mbf{T}^{\text{GT}}\):

\[
\mbf{q}_i^{\text{GT}}
\;=\;
\argmin_{\mbf{q}\in\mbf{T}^{\text{GT}}}
       \bigl\|\mbf{p}_i^{\text{trans}}-\mbf{q}\bigr\|.
\]

The trajectory TRE is then

\[
\text{TRE}_{\text{traj}}
\;=\;
\sqrt{\frac{1}{N_p}\;
      \sum_{i=1}^{N_p}
      \bigl\|
        \mbf{p}_i^{\text{trans}}
        -\mbf{q}_i^{\text{GT}}
      \bigr\|^{2}} .
\]

This nearest-neighbour metric evaluates spatial agreement with the reference trajectory without requiring temporal alignment; it does not quantify tracking latency or along-path lag.

\section{Supplementary Data}\label{sec:additional_results}

\begin{figure*}[width=\textwidth,pos=!tbp]
  \centering
  \begin{subfigure}{0.48\linewidth}
    \centering
    \includegraphics[width=\linewidth]{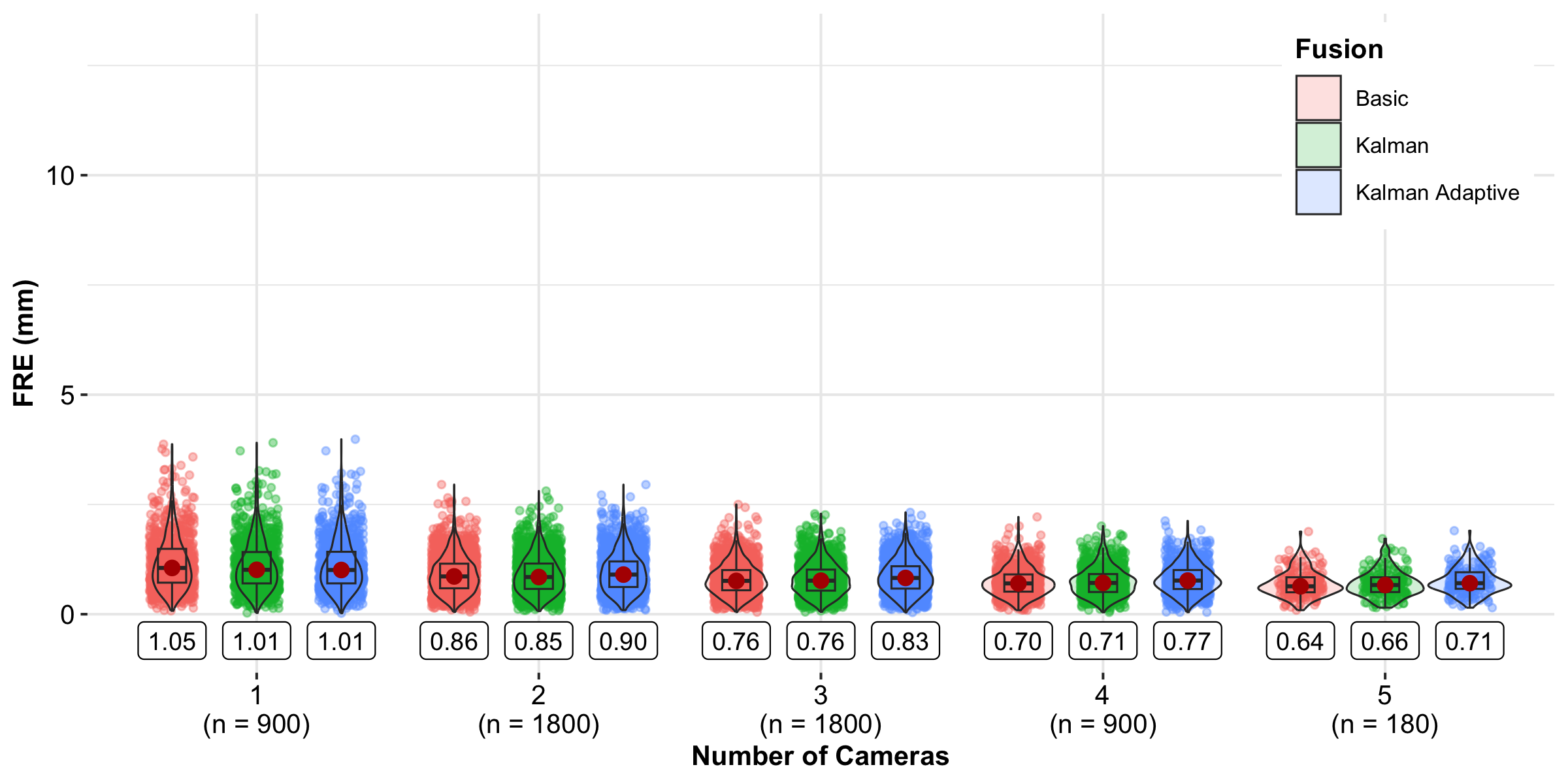}
    \caption{FRE – Monocular Pose Fusion (Grid)}
    \label{fig:fre_grid_mono}
  \end{subfigure}
  \hfill
  \begin{subfigure}{0.48\linewidth}
    \centering
    \includegraphics[width=\linewidth]{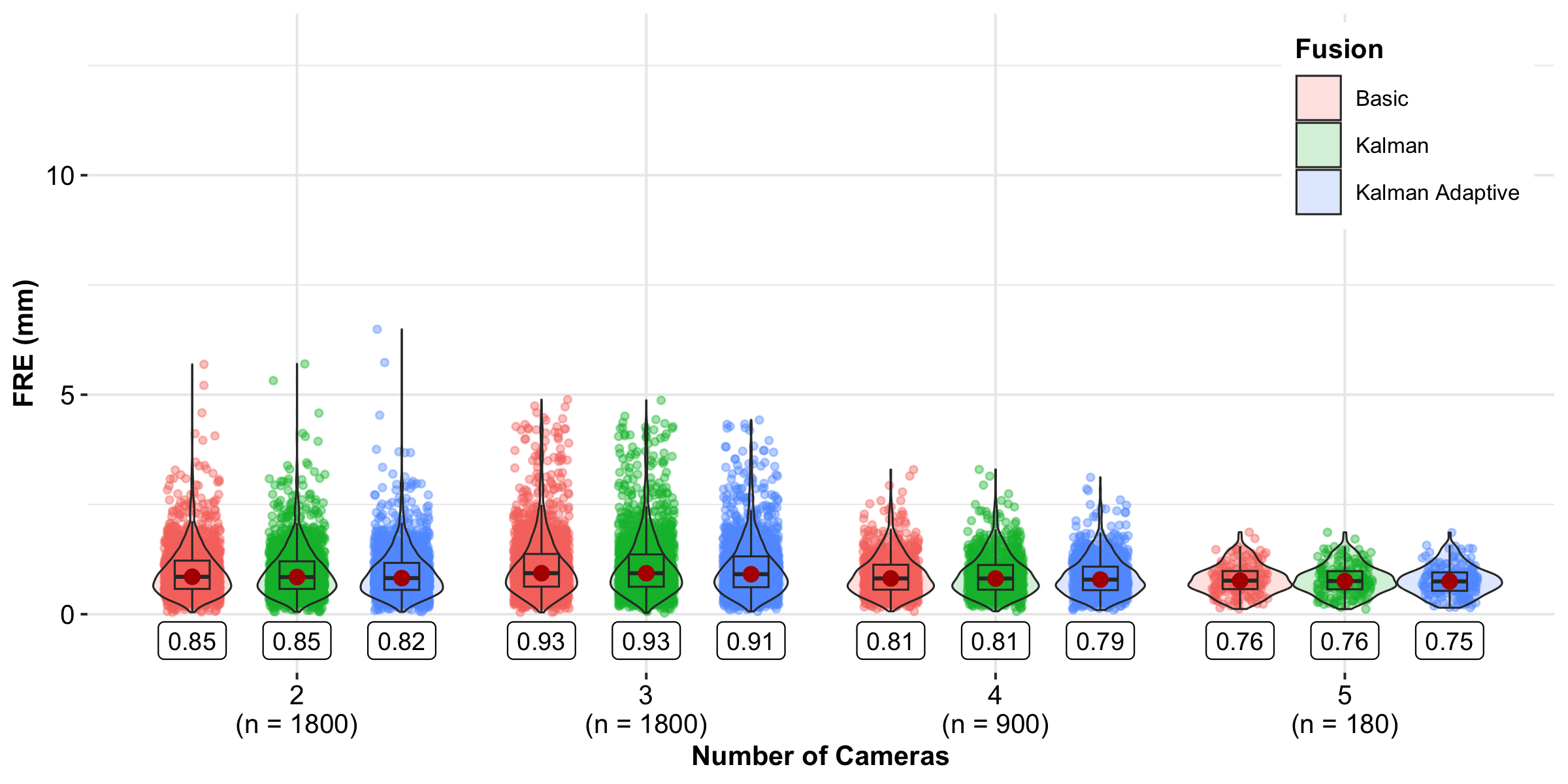}
    \caption{FRE – Triangulation (Grid)}
    \label{fig:fre_grid_triang}
  \end{subfigure}

  \vspace{0.8em} %

  \begin{subfigure}{0.48\linewidth}
    \centering
    \includegraphics[width=\linewidth]{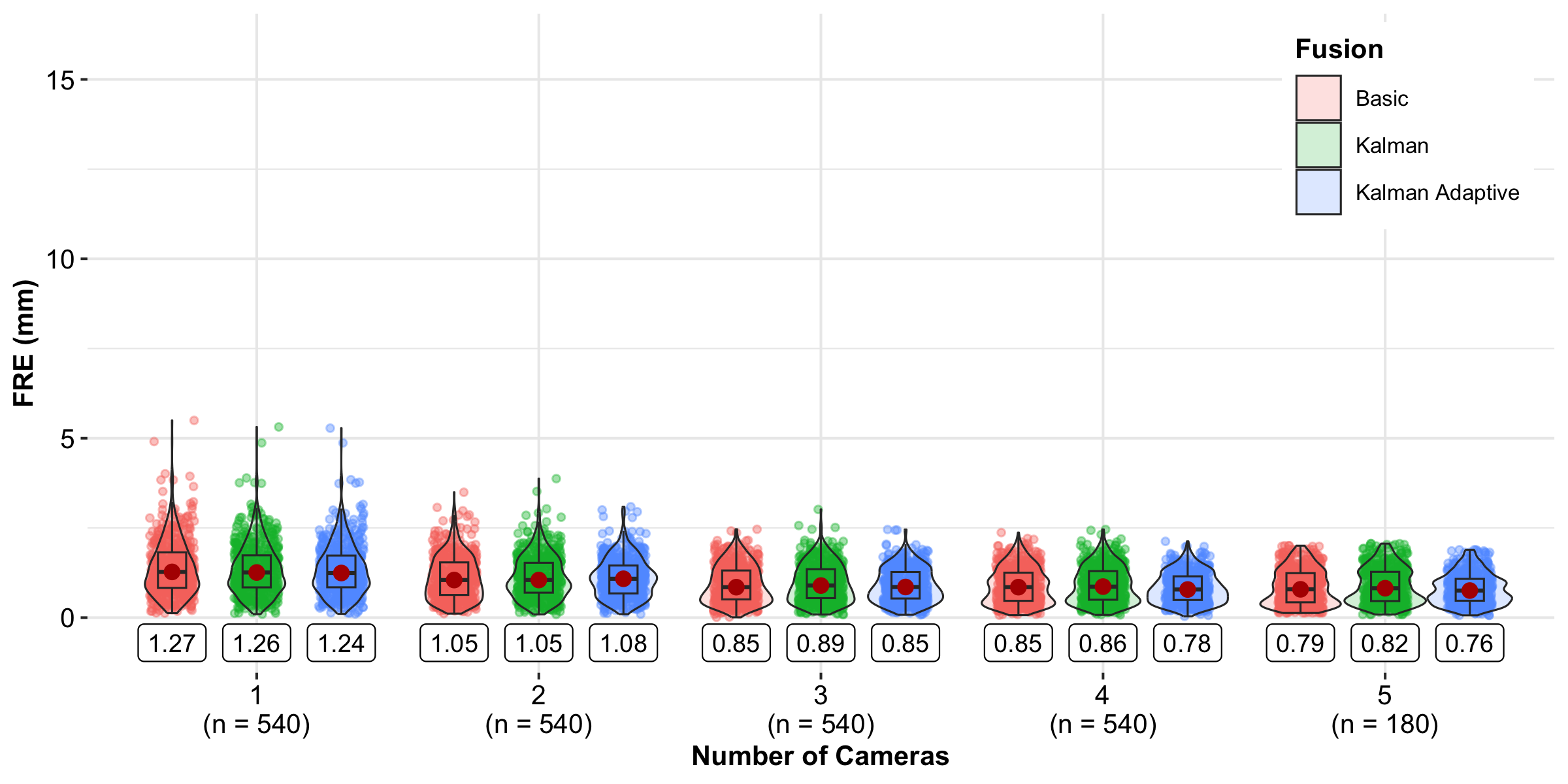}
    \caption{FRE – Monocular Pose Fusion (Head Phantom)}
    \label{fig:fre_head_mono}
  \end{subfigure}
  \hfill
  \begin{subfigure}{0.48\linewidth}
    \centering
    \includegraphics[width=\linewidth]{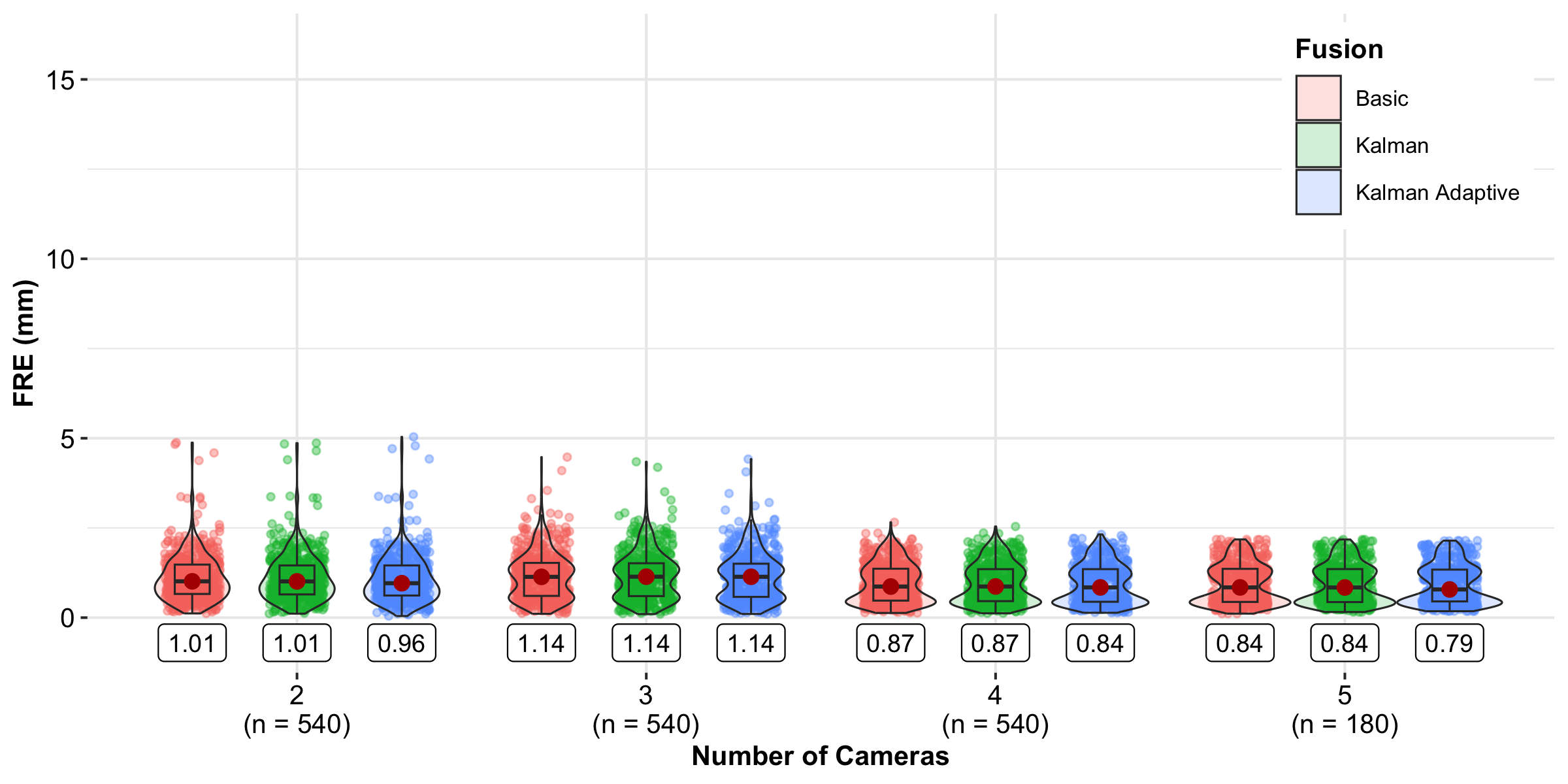}
    \caption{FRE – Triangulation (Head Phantom)}
    \label{fig:fre_head_triang}
  \end{subfigure}

  \caption{Error-distribution violins for FRE across the two tracking pipelines. 
           \textbf{Top row:} Grid experiment; 
           \textbf{Bottom row:} Head Phantom experiment. 
           Left column = Monocular Pose Fusion, Right column = Triangulation. 
           Boxes = IQR, red dots / curved boxes = median, violins = kernel density. Median accuracy value reported in each box below the violin plot. Statistical comparisons use 20 paired trial summaries per phantom.}
  \label{fig:fre_quad}
\end{figure*}

This section provides supplementary data supporting the analyses in the main text. Table~\ref{tab:head_results_combined_occlusion_scenarios} presents detailed per-scenario results for the head phantom experiment, breaking down performance for the top, right, and left occlusion configurations. Figure~\ref{fig:fre_quad} shows FRE distributions for both experiments. Tables~\ref{tab:pairwise_pvals_grid_head}--\ref{tab:pairwise_methods_head} provide complete pairwise statistical comparisons across camera counts and filtering methods.

\begin{table*}[width=\FullWidth,pos=!tbp]
\centering
\caption{Registration RMSE in mm (median ± standard deviation) across 20 Head Phantom Experiment trials for Top, Right, and Left Occlusion Scenarios.}
\resizebox{\textwidth}{!}{
\begin{tabular}{l*{7}{c}}
\toprule
 & \multicolumn{1}{c}{\multirow{2}{*}{\# of Cameras}} & \multicolumn{3}{c}{Monocular Pose Fusion} & \multicolumn{3}{c}{Triangulation} \\
\cmidrule(lr){3-5} \cmidrule(lr){6-8}
 & & Basic & Kalman & Kalman Adaptive & Basic & Kalman & Kalman Adaptive \\
\midrule
\multicolumn{8}{l}{\textbf{Top Occlusion Scenario}} \\
\midrule
\multirow{5}{*}{\makecell{Fiducial Point\\Median Registration RMSE}}
 & 1 & 1.7044 (±0.3059) & 1.6380 (±0.2716) & 1.6414 (±0.2710) & N/A & N/A & N/A \\
 & 2 & 1.3203 (±0.2003) & 1.2829 (±0.1911) & 1.2201 (±0.1589) & 1.1210 (±0.1340) & 1.1065 (±0.1322) & 1.0839 (±0.1245) \\
 & 3 & 1.1006 (±0.1276) & 1.1261 (±0.1387) & 1.1227 (±0.1088) & 1.0796 (±0.1323) & 1.0707 (±0.1310) & 1.0701 (±0.1260) \\
 & 4 & 1.0754 (±0.1302) & 1.0981 (±0.1330) & 0.9978 (±0.0973) & 1.0196 (±0.1273) & 1.0106 (±0.1186) & 1.0093 (±0.0947) \\
 & 5 & 0.9952 (±0.1042) & 1.0332 (±0.1047) & 0.8973 (±0.0977) & 1.0695 (±0.0735) & 1.0686 (±0.0715) & 1.0687 (±0.0649) \\
\cmidrule{1-8}
\multirow{5}{*}{\makecell{Target Point\\Median Registration RMSE}}
 & 1 & 3.1144 (±0.8990) & 3.0852 (±0.8704) & 3.0815 (±0.8676) & N/A & N/A & N/A \\
 & 2 & 1.9501 (±0.6432) & 1.8828 (±0.6241) & 2.0217 (±0.4530) & 1.3348 (±0.4934) & 1.3154 (±0.4601) & 1.2786 (±0.3940) \\
 & 3 & 1.5099 (±0.5814) & 1.5901 (±0.5923) & 1.6486 (±0.4850) & 1.2410 (±0.3526) & 1.2443 (±0.3482) & 1.2410 (±0.3305) \\
 & 4 & 1.6747 (±0.5834) & 1.7824 (±0.5952) & 1.4955 (±0.3396) & 1.5105 (±0.3747) & 1.4975 (±0.3682) & 1.4960 (±0.3562) \\
 & 5 & 1.5759 (±0.4657) & 1.6535 (±0.5066) & 1.4813 (±0.3523) & 1.2997 (±0.1885) & 1.2934 (±0.1897) & 1.2774 (±0.1938) \\
\cmidrule{1-8}
\multirow{5}{*}{\makecell{Target Trajectory\\Median Registration RMSE}}
 & 1 & 2.8075 (±0.6904) & 2.6879 (±0.6730) & 2.6818 (±0.6707) & N/A & N/A & N/A \\
 & 2 & 1.4671 (±0.5825) & 1.3492 (±0.5700) & 1.4175 (±0.4565) & 1.0095 (±0.2685) & 1.0168 (±0.2660) & 0.9843 (±0.2669) \\
 & 3 & 0.9477 (±0.4894) & 1.0509 (±0.5248) & 1.1422 (±0.4320) & 0.9462 (±0.2571) & 0.9396 (±0.2535) & 0.8757 (±0.2505) \\
 & 4 & 1.2358 (±0.4434) & 1.2417 (±0.4728) & 1.2880 (±0.3169) & 1.6845 (±0.5110) & 1.6895 (±0.5013) & 1.6744 (±0.4829) \\
 & 5 & 2.8220 (±1.2112) & 2.6056 (±1.1142) & 1.4894 (±0.3249) & 1.1245 (±0.1549) & 1.1151 (±0.1551) & 1.1197 (±0.1641) \\
\midrule
\multicolumn{8}{l}{\textbf{Right Occlusion Scenario}} \\
\midrule
\multirow{5}{*}{\makecell{Fiducial Point\\Median Registration RMSE}}
 & 1 & 1.4793 (±0.3102) & 1.3591 (±0.2953) & 1.3593 (±0.2951) & N/A & N/A & N/A \\
 & 2 & 1.3039 (±0.2155) & 1.3258 (±0.1876) & 1.2550 (±0.1395) & 1.1168 (±0.1560) & 1.0922 (±0.1625) & 1.0581 (±0.1781) \\
 & 3 & 1.1142 (±0.1380) & 1.1145 (±0.1318) & 0.9859 (±0.1067) & 1.4885 (±0.2799) & 1.4823 (±0.2776) & 1.4742 (±0.2769) \\
 & 4 & 1.0281 (±0.1154) & 1.0401 (±0.1168) & 0.9272 (±0.0980) & 1.1402 (±0.0809) & 1.1341 (±0.0803) & 1.1175 (±0.0797) \\
 & 5 & 0.9952 (±0.1042) & 1.0332 (±0.1047) & 0.8973 (±0.0977) & 1.0695 (±0.0735) & 1.0686 (±0.0715) & 1.0687 (±0.0649) \\
\cmidrule{1-8}
\multirow{5}{*}{\makecell{Target Point\\Median Registration RMSE}}
 & 1 & 3.0655 (±1.7127) & 3.1219 (±1.5448) & 3.1242 (±1.5422) & N/A & N/A & N/A \\
 & 2 & 2.7233 (±0.8803) & 2.7652 (±0.8440) & 2.6429 (±0.4463) & 1.9293 (±0.4214) & 1.9299 (±0.4211) & 1.9409 (±0.4400) \\
 & 3 & 2.1931 (±0.5432) & 2.2858 (±0.5393) & 1.9690 (±0.3273) & 2.4350 (±0.3738) & 2.4673 (±0.3676) & 2.4078 (±0.3623) \\
 & 4 & 1.7375 (±0.4568) & 1.7870 (±0.4484) & 1.7592 (±0.3087) & 1.4504 (±0.2510) & 1.4490 (±0.2551) & 1.4051 (±0.2637) \\
 & 5 & 1.5759 (±0.4657) & 1.6535 (±0.5066) & 1.4813 (±0.3523) & 1.2997 (±0.1885) & 1.2934 (±0.1897) & 1.2774 (±0.1938) \\
\cmidrule{1-8}
\multirow{5}{*}{\makecell{Target Trajectory\\Median Registration RMSE}}
 & 1 & 2.9653 (±0.4777) & 2.9073 (±0.5573) & 2.9148 (±0.5613) & N/A & N/A & N/A \\
 & 2 & 2.4405 (±0.4291) & 2.4779 (±0.4122) & 2.3539 (±0.3399) & 1.9210 (±0.3796) & 1.8870 (±0.3896) & 1.7704 (±0.4092) \\
 & 3 & 4.5664 (±2.2443) & 3.8499 (±1.8934) & 2.0279 (±0.2788) & 2.3899 (±0.4939) & 2.4070 (±0.4932) & 2.2468 (±0.5079) \\
 & 4 & 3.4373 (±1.6122) & 3.1312 (±1.4902) & 1.7619 (±0.2901) & 1.0785 (±0.1689) & 1.0800 (±0.1709) & 1.0534 (±0.1831) \\
 & 5 & 2.8220 (±1.2112) & 2.6056 (±1.1142) & 1.4894 (±0.3249) & 1.1245 (±0.1549) & 1.1151 (±0.1551) & 1.1197 (±0.1641) \\
\midrule
\multicolumn{8}{l}{\textbf{Left Occlusion Scenario}} \\
\midrule
\multirow{5}{*}{\makecell{Fiducial Point\\Median Registration RMSE}}
 & 1 & 1.3240 (±0.3622) & 1.2817 (±0.3480) & 1.2830 (±0.3431) & N/A & N/A & N/A \\
 & 2 & 1.0823 (±0.1837) & 1.1267 (±0.2278) & 1.1479 (±0.2157) & 1.4390 (±0.2801) & 1.4379 (±0.2826) & 1.4375 (±0.2870) \\
 & 3 & 0.9535 (±0.1095) & 0.9849 (±0.1322) & 0.9397 (±0.1345) & 1.2088 (±0.1177) & 1.2100 (±0.1155) & 1.2244 (±0.1085) \\
 & 4 & 0.9884 (±0.0959) & 0.9986 (±0.1153) & 0.8771 (±0.1132) & 1.1107 (±0.0623) & 1.1086 (±0.0627) & 1.1035 (±0.0648) \\
 & 5 & 0.9952 (±0.1042) & 1.0332 (±0.1047) & 0.8973 (±0.0977) & 1.0695 (±0.0735) & 1.0686 (±0.0715) & 1.0687 (±0.0649) \\
\cmidrule{1-8}
\multirow{5}{*}{\makecell{Target Point\\Median Registration RMSE}}
 & 1 & 2.2955 (±0.8513) & 2.1928 (±0.8360) & 2.1547 (±0.8289) & N/A & N/A & N/A \\
 & 2 & 1.7188 (±0.6754) & 1.8659 (±0.7442) & 1.6537 (±0.6140) & 2.0332 (±0.4706) & 2.0251 (±0.4777) & 1.9662 (±0.4912) \\
 & 3 & 1.6418 (±0.5158) & 1.8273 (±0.6123) & 1.5231 (±0.4225) & 1.4532 (±0.2639) & 1.4461 (±0.2580) & 1.4408 (±0.2474) \\
 & 4 & 1.8609 (±0.5333) & 1.8935 (±0.5903) & 1.4831 (±0.3792) & 1.3483 (±0.1689) & 1.3472 (±0.1685) & 1.3647 (±0.1721) \\
 & 5 & 1.5759 (±0.4657) & 1.6535 (±0.5066) & 1.4813 (±0.3523) & 1.2997 (±0.1885) & 1.2934 (±0.1897) & 1.2774 (±0.1938) \\
\cmidrule{1-8}
\multirow{5}{*}{\makecell{Target Trajectory\\Median Registration RMSE}}
 & 1 & 1.5482 (±0.7737) & 1.4661 (±0.7600) & 1.4502 (±0.7450) & N/A & N/A & N/A \\
 & 2 & 1.3190 (±0.4893) & 1.3271 (±0.6025) & 1.2349 (±0.4687) & 1.3880 (±0.3130) & 1.3727 (±0.3154) & 1.2811 (±0.3248) \\
 & 3 & 4.4467 (±2.3635) & 3.7694 (±2.0250) & 1.4716 (±0.4236) & 1.4905 (±0.2860) & 1.4785 (±0.2856) & 1.4609 (±0.2918) \\
 & 4 & 3.4916 (±1.5786) & 3.1765 (±1.4025) & 1.5682 (±0.3566) & 1.2017 (±0.1898) & 1.1969 (±0.1919) & 1.1478 (±0.2060) \\
 & 5 & 2.8220 (±1.2112) & 2.6056 (±1.1142) & 1.4894 (±0.3249) & 1.1245 (±0.1549) & 1.1151 (±0.1551) & 1.1197 (±0.1641) \\
\bottomrule
\end{tabular}
}
\label{tab:head_results_combined_occlusion_scenarios}
\end{table*}

\begin{table*}[width=\FullWidth,pos=!tbp]
\centering
\caption{Two-sided paired Wilcoxon signed-rank $p$-values for pooled camera-count comparisons, with Holm correction over the ten count pairs within each experiment and metric. Trial mean squares are averaged equally across available tracking methods and filters before taking the square root (pose fusion only at one camera; both tracking methods at two to five cameras). All tests use 20 paired trial summaries. $\dagger$ marks a comparison whose significance decision at $\alpha=0.05$ differs under the Holm-adjusted paired sign-test sensitivity analysis.}
\resizebox{\textwidth}{!}{
\begin{tabular}{c*{6}{r}}
\toprule
\multirow{2}{*}{Camera counts} & \multicolumn{3}{c}{Grid} & \multicolumn{3}{c}{Head phantom} \\
\cmidrule(lr){2-4} \cmidrule(lr){5-7}
 & FRE & Point TRE & Trajectory TRE & FRE & Point TRE & Trajectory TRE \\
\midrule
1 versus 2 & 1.91e-05 & 1.91e-05 & 3.15e-04 & 1.91e-05 & 1.91e-05 & 1.91e-05 \\
1 versus 3 & 2.67e-05 & 1.91e-05 & 0.0408$^{\dagger}$ & 2.86e-05 & 1.91e-05 & 0.8182 \\
1 versus 4 & 1.91e-05 & 1.91e-05 & 0.1392 & 1.91e-05 & 1.91e-05 & 0.0099$^{\dagger}$ \\
1 versus 5 & 1.91e-05 & 1.91e-05 & 1.14e-04 & 1.91e-05 & 1.91e-05 & 0.0029 \\
2 versus 3 & 0.0136 & 0.0073 & 1.91e-05 & 0.0032 & 1.34e-04 & 0.0099$^{\dagger}$ \\
2 versus 4 & 1.91e-05 & 1.91e-05 & 0.4091 & 1.91e-05 & 1.91e-05 & 0.0983 \\
2 versus 5 & 1.91e-05 & 1.91e-05 & 5.34e-04 & 1.91e-05 & 1.91e-05 & 0.8182 \\
3 versus 4 & 1.91e-05 & 1.91e-05 & 1.91e-05 & 1.91e-05 & 1.91e-05 & 0.0012 \\
3 versus 5 & 1.91e-05 & 1.91e-05 & 1.91e-05 & 1.91e-05 & 1.91e-05 & 1.91e-05 \\
4 versus 5 & 1.91e-05 & 1.91e-05 & 1.91e-05 & 0.0020 & 1.91e-05 & 1.91e-05 \\
\bottomrule
\end{tabular}
}
\label{tab:pairwise_pvals_grid_head}
\end{table*}

\begin{table*}[width=\FullWidth,pos=!tbp]
\centering
\caption{Two-sided paired Wilcoxon signed-rank $p$-values for filtering methods (Grid), with Holm correction over the three filter pairs within each camera count, tracking method, and metric. All tests use 20 paired trial summaries. $\dagger$ marks a comparison whose significance decision at $\alpha=0.05$ differs under the Holm-adjusted paired sign-test sensitivity analysis.}
\resizebox{\textwidth}{!}{
\begin{tabular}{c c c r r r}
\toprule
Camera count & Tracking method & Metric & Basic versus Kalman & Basic versus Kalman Adaptive & Kalman versus Kalman Adaptive \\
\midrule
\multirow{3}{*}{1} & \multirow{3}{*}{\makecell{Monocular \\ Pose Fusion}} & FRE & 1.89e-04 & 1.89e-04 & 0.8124 \\
 &  & Point TRE & 5.04e-04 & 9.65e-04 & 0.7562 \\
 &  & Trajectory TRE & 5.72e-06 & 5.72e-06 & 1.05e-04 \\
\midrule
\multirow{3}{*}{2} & \multirow{3}{*}{\makecell{Monocular \\ Pose Fusion}} & FRE & 0.5459 & 0.0097$^{\dagger}$ & 0.0060$^{\dagger}$ \\
 &  & Point TRE & 0.2858 & 0.2858 & 0.1595 \\
 &  & Trajectory TRE & 5.72e-06 & 1.14e-05 & 0.0192 \\
\midrule
\multirow{3}{*}{2} & \multirow{3}{*}{Triangulation} & FRE & 5.72e-06 & 5.72e-06 & 5.72e-06 \\
 &  & Point TRE & 5.72e-06 & 5.72e-06 & 5.72e-06 \\
 &  & Trajectory TRE & 5.72e-06 & 5.72e-06 & 5.72e-06 \\
\midrule
\multirow{3}{*}{3} & \multirow{3}{*}{\makecell{Monocular \\ Pose Fusion}} & FRE & 0.5459 & 0.0026 & 0.0029 \\
 &  & Point TRE & 0.9563 & 0.3162 & 0.3162 \\
 &  & Trajectory TRE & 1.72e-05 & 1.64e-04 & 0.0532 \\
\midrule
\multirow{3}{*}{3} & \multirow{3}{*}{Triangulation} & FRE & 2.86e-05 & 2.86e-05 & 2.86e-05 \\
 &  & Point TRE & 5.72e-06 & 5.72e-06 & 5.72e-06 \\
 &  & Trajectory TRE & 5.72e-06 & 5.72e-06 & 5.72e-06 \\
\midrule
\multirow{3}{*}{4} & \multirow{3}{*}{\makecell{Monocular \\ Pose Fusion}} & FRE & 0.4524 & 0.0081 & 0.0166$^{\dagger}$ \\
 &  & Point TRE & 0.7768 & 0.6485 & 0.7768 \\
 &  & Trajectory TRE & 3.15e-04 & 3.36e-04 & 0.1054 \\
\midrule
\multirow{3}{*}{4} & \multirow{3}{*}{Triangulation} & FRE & 1.34e-04 & 1.09e-04 & 1.09e-04 \\
 &  & Point TRE & 5.72e-06 & 1.91e-05 & 1.91e-05 \\
 &  & Trajectory TRE & 5.72e-06 & 5.72e-06 & 5.72e-06 \\
\midrule
\multirow{3}{*}{5} & \multirow{3}{*}{\makecell{Monocular \\ Pose Fusion}} & FRE & 0.3884 & 0.0515 & 0.0515 \\
 &  & Point TRE & 1.0000 & 1.0000 & 1.0000 \\
 &  & Trajectory TRE & 5.04e-04 & 7.90e-04 & 0.1327 \\
\midrule
\multirow{3}{*}{5} & \multirow{3}{*}{Triangulation} & FRE & 0.0110 & 0.0112$^{\dagger}$ & 0.0112$^{\dagger}$ \\
 &  & Point TRE & 1.72e-05 & 1.26e-04 & 1.34e-04 \\
 &  & Trajectory TRE & 5.72e-06 & 5.72e-06 & 5.72e-06 \\
\bottomrule
\end{tabular}
}
\label{tab:pairwise_methods_grid}
\end{table*}

\begin{table*}[width=\FullWidth,pos=!tbp]
\centering
\caption{Two-sided paired Wilcoxon signed-rank $p$-values for filtering methods (Head phantom), with Holm correction over the three filter pairs within each camera count, tracking method, and metric. All tests use 20 paired trial summaries. $\dagger$ marks a comparison whose significance decision at $\alpha=0.05$ differs under the Holm-adjusted paired sign-test sensitivity analysis.}
\resizebox{\textwidth}{!}{
\begin{tabular}{c c c r r r}
\toprule
Camera count & Tracking method & Metric & Basic versus Kalman & Basic versus Kalman Adaptive & Kalman versus Kalman Adaptive \\
\midrule
\multirow{3}{*}{1} & \multirow{3}{*}{\makecell{Monocular \\ Pose Fusion}} & FRE & 7.63e-06 & 5.72e-06 & 8.51e-04 \\
 &  & Point TRE & 0.4049 & 0.4049 & 0.0983 \\
 &  & Trajectory TRE & 0.0592 & 0.0577 & 0.0592 \\
\midrule
\multirow{3}{*}{2} & \multirow{3}{*}{\makecell{Monocular \\ Pose Fusion}} & FRE & 0.6976 & 0.6976 & 0.4949 \\
 &  & Point TRE & 0.1794 & 0.2611 & 0.0719 \\
 &  & Trajectory TRE & 0.8124 & 0.0645 & 0.1165 \\
\midrule
\multirow{3}{*}{2} & \multirow{3}{*}{Triangulation} & FRE & 0.0017 & 0.0012 & 0.0012 \\
 &  & Point TRE & 0.0242$^{\dagger}$ & 0.0242$^{\dagger}$ & 0.0219 \\
 &  & Trajectory TRE & 0.0023 & 1.14e-05 & 5.72e-06 \\
\midrule
\multirow{3}{*}{3} & \multirow{3}{*}{\makecell{Monocular \\ Pose Fusion}} & FRE & 0.0592$^{\dagger}$ & 0.0592 & 0.0219$^{\dagger}$ \\
 &  & Point TRE & 0.0242$^{\dagger}$ & 0.0242$^{\dagger}$ & 0.0070$^{\dagger}$ \\
 &  & Trajectory TRE & 0.0037 & 1.64e-04 & 1.09e-04 \\
\midrule
\multirow{3}{*}{3} & \multirow{3}{*}{Triangulation} & FRE & 0.0020 & 4.01e-04 & 4.01e-04 \\
 &  & Point TRE & 0.0095 & 0.0095 & 0.0095 \\
 &  & Trajectory TRE & 8.51e-04 & 7.63e-06 & 5.72e-06 \\
\midrule
\multirow{3}{*}{4} & \multirow{3}{*}{\makecell{Monocular \\ Pose Fusion}} & FRE & 0.0083 & 0.0034 & 4.01e-04 \\
 &  & Point TRE & 0.0121 & 0.0047 & 0.0012 \\
 &  & Trajectory TRE & 0.0032$^{\dagger}$ & 3.36e-04 & 1.43e-04 \\
\midrule
\multirow{3}{*}{4} & \multirow{3}{*}{Triangulation} & FRE & 0.0018 & 0.0018 & 0.0018 \\
 &  & Point TRE & 0.0321$^{\dagger}$ & 0.0321$^{\dagger}$ & 0.0321$^{\dagger}$ \\
 &  & Trajectory TRE & 0.3884 & 0.0242$^{\dagger}$ & 0.0051$^{\dagger}$ \\
\midrule
\multirow{3}{*}{5} & \multirow{3}{*}{\makecell{Monocular \\ Pose Fusion}} & FRE & 0.0136$^{\dagger}$ & 0.0084$^{\dagger}$ & 5.04e-04 \\
 &  & Point TRE & 0.1429 & 0.0242$^{\dagger}$ & 0.0219$^{\dagger}$ \\
 &  & Trajectory TRE & 0.0014 & 4.01e-04 & 4.01e-04 \\
\midrule
\multirow{3}{*}{5} & \multirow{3}{*}{Triangulation} & FRE & 0.0219$^{\dagger}$ & 0.0385$^{\dagger}$ & 0.0385$^{\dagger}$ \\
 &  & Point TRE & 0.0515 & 0.0515 & 0.0515 \\
 &  & Trajectory TRE & 0.3683 & 0.1063 & 0.0577 \\
\bottomrule
\end{tabular}
}
\label{tab:pairwise_methods_head}
\end{table*}

\clearpage

\clearpage

\section*{Declaration of generative AI and AI-assisted technologies in the manuscript preparation process}
During the preparation of this work the authors used Claude (Anthropic) in order to assist with language editing, \LaTeX{} formatting, and code review and revision. After using this tool, the authors reviewed and edited the content as needed and take full responsibility for the content of the published article.

\section*{Data availability}
The hardware designs, technical drawings, bill of materials, and the calibration, tracking, and registration software are available at \url{https://nist.mni.mcgill.ca/software/}. The 5,000 synchronized multi-view tracking videos, the simultaneously recorded fusionTrack 500 measurements, and the per-trial registration-error results are available at \url{https://nist.mni.mcgill.ca/data/}. All materials are released under the Creative Commons Attribution 4.0 (CC BY 4.0) licence.

\printcredits

\bibliographystyle{cas-model2-names}
\bibliography{references}

\end{document}